%% file: main.tex
\documentclass[10pt,twocolumn,letterpaper,table]{article}

\usepackage[pagenumbers]{cvpr} 

\usepackage{xcolor}
\definecolor{cvprblue}{rgb}{0.21,0.49,0.74}
\usepackage[pagebackref=true,breaklinks=true,colorlinks,citecolor=cvprblue,linkcolor=cvprblue,bookmarks=false]{hyperref}
\usepackage[accsupp]{axessibility}  
\usepackage{graphicx}
\usepackage{url}
\usepackage{booktabs}
\usepackage{multirow}
\usepackage{pifont}
\usepackage{xspace}
\usepackage{makecell}
\usepackage{enumitem}
\setitemize{noitemsep,topsep=0pt,parsep=0pt,partopsep=0pt}
\newcommand{\mycaption}[2]{\caption{\textbf{#1.}~#2}}
\newcommand{\cmark}{\ding{51}}
\newcommand{\xmark}{\ding{55}}
\newcommand{\cellfirst}{\cellcolor{Green!50}}
\newcommand{\cellsecond}{\cellcolor{LimeGreen!25}}
\newcommand\model{\emph{Video Alchemist}\xspace}
\newcommand\benchmark{\emph{MSRVTT-Personalization}\xspace}
\newcommand\blfootnote[1]{
  \begingroup
  \renewcommand\thefootnote{}\footnote{#1}
  \addtocounter{footnote}{-1}
  \endgroup
}
\newcommand{\myparagraph}[1]{\noindent\textbf{#1}}

\def\paperID{5792} 
\def\confName{CVPR}
\def\confYear{2025}

\title{Multi-subject Open-set Personalization in Video Generation}

\author{
Tsai-Shien Chen$^{1,2,*}$ \quad Aliaksandr Siarohin$^1$ \quad Willi Menapace$^1$ \quad Yuwei Fang$^1$ 
\quad Kwot Sin Lee$^1$ \\ Ivan Skorokhodov$^1$ \quad Kfir Aberman$^1$ \quad Jun-Yan Zhu$^{3}$ \quad Ming-Hsuan Yang$^{2}$ \quad Sergey Tulyakov$^1$ \\
{\normalsize $^1$Snap Inc. \quad $^2$UC Merced \quad $^3$CMU} \\
{\small \url{https://snap-research.github.io/open-set-video-personalization}}
}

\begin{document}

\input{figures/teaser}

\blfootnote{\hspace{-.6cm} $^*$This work was done while interning at Snap.}

\iftoggle{arxiv}{\clearpage}{}
\input{sources/0_abstract}
\input{sources/1_introduction}
\input{sources/2_related_work}
\input{sources/3_methodology}
\input{sources/4_experiment}
\input{sources/5_conclusion}

{
    \small
    \bibliographystyle{ieeenat_fullname}
    \bibliography{main}
}

\appendix
\maketitlesupplementary

\input{sources/A_detail_dataset}
\input{sources/B_detail_model}
\input{sources/C_visualization}
\clearpage
\input{sources/D_limitations}

\end{document}

%% file: figures/teaser.tex
\twocolumn[{
\renewcommand\twocolumn[1][]{#1}
\maketitle
\begin{center}
    \centering
    \captionsetup{type=figure}
    \iftoggle{arxiv}{}{\vspace{-5.5mm}}
    \includegraphics[width=\textwidth]{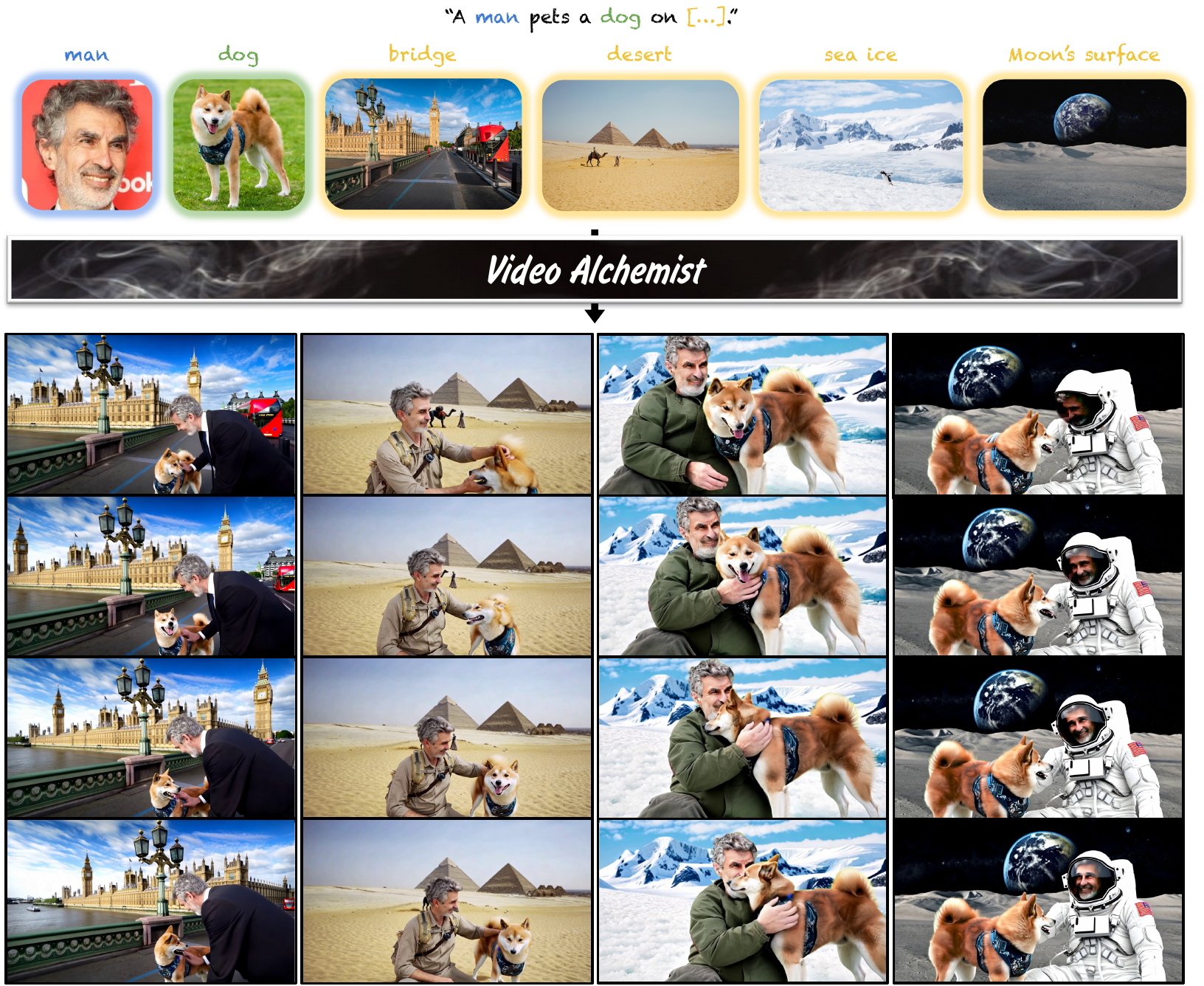}
    \iftoggle{arxiv}{\vspace{-4mm}}{\vspace{-6.5mm}}
    \caption{
        Given a text prompt as well as reference images for each subject (\ie, \textit{man}, \textit{dog}) and background images (\ie, \textit{bridge}, \textit{desert}, \textit{sea ice}, \textit{Moon's surface}), \model synthesizes natural motions while preserving subject identity and background fidelity. 
    }
    \label{fig:teaser}
\end{center}
}]

%% file: sources/0_abstract.tex
\iftoggle{arxiv}{}{\vspace{-3mm}}
\begin{abstract}
\vspace{-7.5mm}

Video personalization methods allow us to synthesize videos with specific concepts such as people, pets, and places. However, existing methods often focus on limited domains, require time-consuming optimization per subject, or support only a single subject. We present Video Alchemist---a video model with built-in multi-subject, open-set personalization capabilities for both foreground objects and background, eliminating the need for time-consuming test-time optimization. Our model is built on a new Diffusion Transformer module that fuses each conditional reference image and its corresponding subject-level text prompt with cross-attention layers. Developing such a large model presents two main challenges: dataset and evaluation. First, as paired datasets of reference images and videos are extremely hard to collect, we sample selected video frames as reference images and synthesize a clip of the target video. However, while models can easily denoise training videos given reference frames, they fail to generalize to new contexts. To mitigate this issue, we design a new automatic data construction pipeline with extensive image augmentations. Second, evaluating open-set video personalization is a challenge in itself. To address this, we introduce a personalization benchmark that focuses on accurate subject fidelity and supports diverse personalization scenarios. Finally, our extensive experiments show that our method significantly outperforms existing personalization methods in both quantitative and qualitative evaluations. 

\end{abstract}

%% file: sources/1_introduction.tex
\iftoggle{arxiv}{\vspace{-5mm}}{\vspace{-6.5mm}}
\section{Introduction}
\label{sec:introduction}
\iftoggle{arxiv}{}{\vspace{-1.5mm}}

Diffusion models~\cite{ddpm,sohl2015deep,song2019generative} have enabled us to synthesize realistic videos with natural motions from text prompts~\cite{make_a_video,videoldm,sora,vdm,snapvideo}. This level of quality and realism paves the way for personalization---the ability to generate videos containing specific objects and people in unseen contexts or backgrounds. Multiple methods have been proposed to generate content with specific people or pets, but they remain limited to closed-set object categories. Some only support human faces~\cite{id_animator,magicme} or a single subject~\cite{videobooth,dreamvideo,storydiffusion,customcrafter}, while others only work on foreground objects~\cite{videodreamer,customvideo,disenstudio}. Moreover, many of these methods require costly test-time optimization~\cite{dreamvideo,customcrafter,magicme}.

In this paper, we present \model, a video generation model with extensive personalization capabilities. Our model supports the customization of multiple subjects and open-set entities, including both foreground objects and background. Importantly, our method does not require fine-tuning to incorporate new concepts. Figure~\ref{fig:teaser} shows videos personalized for two subjects across four backgrounds. \model is built on new Diffusion Transformer modules~\cite{dit} tailored for personalization. Each module uses two cross-attention layers: one to integrate the text prompt describing the entire video and the other to incorporate the embeddings of each reference image. To achieve multi-subject conditioning, we employ a simple yet effective subject-level fusion, blending the word description of each subject with its image embeddings.

But how can we collect data to train our model? Ideally, it requires a dataset of videos and images with many subjects, each captured with varying lighting, background, and pose. Unfortunately, collecting such a dataset for open-set entities is challenging at best and impossible at worst. Alternatively, we can extract reference images and target video clips from the same video. However, this approach comes with a significant drawback---factors unrelated to identity still have a very high correlation across different video frames, leading to what we term the \textit{copy-and-paste} effect. This issue is commonly seen in reconstruction-based methods, such as IP-Adapter~\cite{ip_adapter}, as shown in Figure~\ref{fig:msrvtt_comparison}. As a result, the model struggles to synthesize diverse videos with unseen backgrounds, lighting, and pose. To alleviate this overfitting, we design a data construction pipeline to automatically extract object segments from target videos and craft personalization-specific data augmentation to ensure that the model focuses on the subject identity of the reference images. Experiments show that training with the proposed augmentation can significantly mitigate the \textit{copy-and-paste} effect, as shown in Figure~\ref{fig:ablation_study}.

Another challenge is the lack of a suitable benchmark for evaluating multi-subject video personalization. Typically, we evaluate video personalization results by computing a similarity score between the generated video and the reference images~\cite{dreambooth,ip_adapter,videobooth,storydiffusion}. Unfortunately, this metric does not apply to multiple entities, as it cannot focus on each subject separately. To address these limitations, we introduce \benchmark, a comprehensive and robust evaluation protocol for personalization tasks. This new benchmark facilitates evaluation across various conditioning modes, including conditioning on face crops, single or multiple arbitrary subjects, and combinations of foreground objects and backgrounds. Unlike image-level similarity, we evaluate the subject fidelity of each object segment. The experiments demonstrate that \model outperforms existing personalization methods regarding both quantitative and qualitative evaluations. In addition, we conduct an extensive ablation study to verify the effectiveness of our proposed components. 

\input{figures/dataset_collection}

Our contributions can be summarized as follows:
\begin{itemize}[leftmargin=8pt]
    \item We present \model, a new video generation model that supports multi-subject, open-set personalization for both foreground objects and background.
    \item We carefully curate a large-scale training dataset and introduce training techniques to reduce model overfitting.
    \item We introduce \benchmark, a new video personalization benchmark, providing various conditioning modes and accurate measurement of subject fidelity.
\end{itemize}

%% file: figures/dataset_collection.tex
\begin{figure*}[t]
    \centering
    \includegraphics[width=\linewidth]{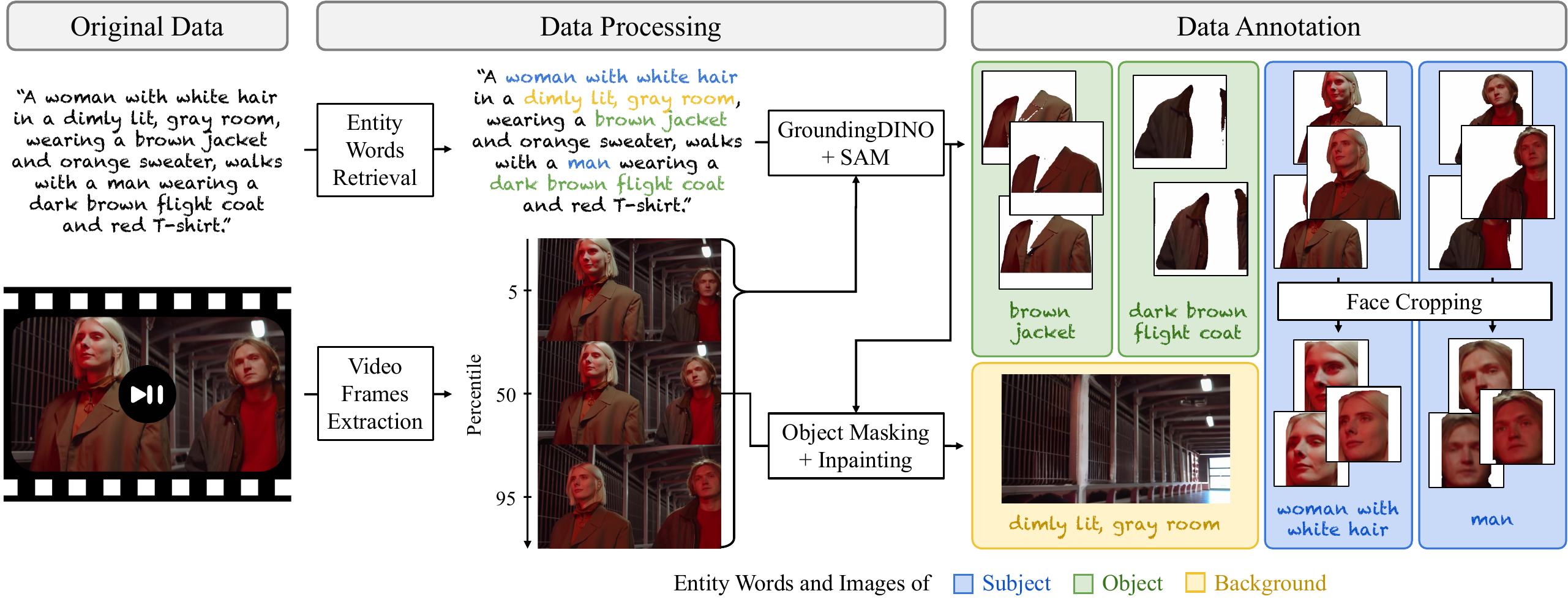}
    \iftoggle{arxiv}{\vspace{-5mm}}{\vspace{-8.5mm}}
    \mycaption{Dataset collection pipeline for video personalization}{
        We construct our training dataset using video and caption pairs through three steps.  First, we identify three categories of entity words from the caption: subject, object, and background. Next, we use these entity words to localize and segment the target subjects and objects in three selected video frames. Finally, we extract a clean background image by removing the subjects and objects from the middle frame.
    }
    \label{fig:dataset_collecition}
    \iftoggle{arxiv}{\vspace{+2mm}}{\vspace{-5mm}}
\end{figure*}

%% file: sources/2_related_work.tex
\iftoggle{arxiv}{}{\vspace{-2mm}}
\section{Related Work}
\label{sec:related_work}
\iftoggle{arxiv}{}{\vspace{-1.5mm}}

\myparagraph{Diffusion Video Models.}
Diffusion models~\cite{sohl2015deep,song2019generative,ddpm,ldm,vdm} have demonstrated impressive capabilities in generating realistic images. Building on this success, recent studies have explored their applications in text-conditioned video synthesis~\cite{imagen,make_a_video,videoldm,magicvideo,videofusion,animatediff,mahapatra2023text,girdhar2023emu,cogvideox,snapvideo,sora}. ImagenVideo~\cite{imagen} and Make-A-Video~\cite{make_a_video} use cascaded temporal and spatial upsamplers for video generation. VideoLDM~\cite{videoldm} fine-tunes a pre-trained latent image generator and decoder to produce temporally coherent videos. Differently from previous models based on the U-Net~\cite{unet} architecture, SnapVideo~\cite{snapvideo} adapts the FiT~\cite{fit} and scales to billion-parameter models. More recently, Sora~\cite{sora} adopts the Diffusion Transformer~\cite{dit} to achieve high-resolution, long video synthesis. While these studies have shown significant progress, using text prompts alone confines the generated content to what can be described in words.

\iftoggle{arxiv}{\vspace{2mm}}{}
\myparagraph{Personalized Image Generation.}
This task aims to customize a generative model to new concepts and subjects using a few input images~\cite{dreambooth,textual_inversion,multi_concept_customization,han2023svdiff,ip_adapter,instantbooth,consistory,instantid,moa,avrahami2023break,jones2024customizing,voynov2023p+,ham2024personalized}. For example, DreamBooth~\cite{dreambooth} optimizes the entire text-to-image model for each subject, while Textual inversion~\cite{textual_inversion} learns a text embedding for each subject and uses the embedding to generate novel images. Custom Diffusion~\cite{multi_concept_customization} learns to compose multiple concepts, each represented by text embedding and cross-attention weights. However, these optimization-based models require finetuning weights or optimizing embeddings for every new concept, which is inevitably slow and prone to overfitting. 

Recent studies have explored encoder-based methods to reduce test-time finetuning~\cite{instantbooth,ip_adapter,arar2023domain,gal2023encoder,elite,li2023blip,chen2024anydoor,xiao2024fastcomposer,valevski2023face0,hyperdreambooth}. IP-adapter~\cite{ip_adapter} learns a lightweight decoupled cross-attention mechanism for image conditioning. InstanceBooth~\cite{instantbooth} trains an image encoder to convert reference images into textual tokens and introduces adapter layers to retain identity details. Our model also uses an encoder, but we focus on video personalization with multiple subjects. 

\iftoggle{arxiv}{\vspace{2mm}}{}
\myparagraph{Personalized Video Generation.}
Several works have extended model personalization techniques for videos~\cite{moonshot,videobooth,dreamvideo,customvideo,videodrafter,mcdiff,storydiffusion,materzynska2023customizing,customcrafter,id_animator,magicme,vimi}. DreamVideo~\cite{dreamvideo} uses an optimization-based strategy, training an image adapter to capture the subject's appearance and a motion adapter to model dynamics. In contrast, StoryDiffusion~\cite{storydiffusion} adopts an optimization-free approach with a consistent self-attention mechanism and a semantic motion predictor to ensure smooth transitions and consistent subjects. 

However, most of the existing methods focus on limited domains. Some are limited to face personalization~\cite{id_animator,magicme} or a single subject from specific categories~\cite{moonshot,videobooth,dreamvideo,storydiffusion,customcrafter}, while others focus solely on foreground objects~\cite{videodreamer,customvideo,disenstudio}. In contrast, we introduce a video model that supports the customization of multiple open-set entities across both foreground objects and background. Closely related to our work, VideoDrafter~\cite{videodrafter} achieves open-set video personalization in two stages: text-to-image personalization and first-frame animation. In contrast, our end-to-end method alleviates poor subject consistency in long video synthesis, a notable limitation of first-frame animation.

%% file: sources/3_methodology.tex
\input{figures/model_architecture}

\iftoggle{arxiv}{}{\vspace{-2mm}}
\section{Methodology}
\label{sec:method}
\iftoggle{arxiv}{}{\vspace{-1.5mm}}

Our goal is to learn a generative video model conditioned on a text prompt and a set of images representing each entity word in the prompt.

\iftoggle{arxiv}{}{\vspace{-1.5mm}}
\subsection{Dataset Collection}
\label{sec:method_dataset}
\iftoggle{arxiv}{}{\vspace{-1mm}}

As shown in Figure~\ref{fig:dataset_collecition}, we curate our dataset in three steps.

\iftoggle{arxiv}{\vspace{2mm}}{}
\myparagraph{Retrieving Entity Words.}
To achieve multi-subject personalization, we use a large language model~\cite{mistral} to retrieve multiple entity words from a single caption. Specifically, we define three types of entity words: subject (\eg, human, animal), object (\eg, car, jacket), and background (\eg, room, beach). Subjects and objects are supposed to be clearly visible in the video. Next, we adapt several criteria to filter and enhance the quality of the training dataset. For example,  we exclude videos with any subject entity word in plural form (\eg, a group of people, several dogs) to avoid ambiguity in personalization. Another example is that we remove videos without subject entity words, as their dynamics is often dominated by meaningless camera movements. More details can be found in Appendix~\ref{app:details_datasets_retrieval}.

\iftoggle{arxiv}{\vspace{2mm}}{}
\myparagraph{Preparing Subject Images.}
Next, we select three frames from the beginning, middle, and end of the video (in the 5\%, 50\%, and 95\% percentiles). The motivation is to capture the target subject or object with different poses and lighting conditions. Subsequently, we apply GroundingDINO~\cite{grounding_dino} to each frame to detect the bounding boxes. These bounding boxes are then used by SAM~\cite{sam} to segment the mask regions corresponding to each entity. Moreover, for reference images depicting humans, we apply face detection~\cite{yolov9} to extract face crops.

\iftoggle{arxiv}{\vspace{2mm}}{}
\myparagraph{Preparing Background Image.}
Lastly, we create a clean background image by removing the subjects and objects. Since SAM~\cite{sam} occasionally produces imprecise boundaries, we dilate the foreground mask before applying an inpainting algorithm~\cite{ldm}. We use the background entity word as the positive prompt and use ``\textit{Any human or any object, complex pattern, and texture}'' as the \emph{negative} prompt. To ensure background consistency, we only use the middle frame of each video sequence.

\iftoggle{arxiv}{}{\vspace{-1.5mm}}
\subsection{Video Personalization Model}
\label{sec:method_framework}
\iftoggle{arxiv}{}{\vspace{-1mm}}

We learn \model by denoising the video using a text prompt, reference images, and their corresponding entity words as conditions.

\iftoggle{arxiv}{\vspace{2mm}}{}
\myparagraph{Video Generation Backbone.}
As illustrated in Figure~\ref{fig:model_architecture}, our model is a latent Diffusion Transformer (DiT)~\cite{dit}, where we first compress a video into a latent representation using an autoencoder~\cite{cogvideox} and encode it into a sequence of 1-D video tokens with a tokenizer~\cite{tokenizer}. Next, we add Gaussian noise to obtain a noisy sample and learn a denoising network following the rectified flow formulation~\cite{cnf,rfflow}.

Our network is a deep cascade of DiT blocks. Unlike vanilla DiT designs, our module supports built-in personalization capability by combining information from both text and image conditioning. Our DiT block includes three layers: one multi-head self-attention~\cite{transformer}, followed by two multi-head cross-attention for text and personalization conditioning, respectively. We use RoPE~\cite{rope} positional embeddings in self-attention due to its effectiveness irrespective of number of video tokens and use self-conditioning~\cite{self_conditioning} to enhance visual quality. We further adopt flash attention~\cite{flash_attention} and the fused layer norm~\cite{fused_layer_norm} to accelerate the model training and inference.

\iftoggle{arxiv}{\vspace{2mm}}{}
\myparagraph{Binding of Image and Word Concepts.}
For multi-subject personalization, the model can be conditioned on different subjects, each represented by one or more reference images. Consequently, providing the binding between corresponding text tokens and image tokens is critical. As shown in the second row of Figure~\ref{fig:ablation_study}, without such binding information, the model tends to apply image conditioning to an incorrect subject, such as placing a reference human face on a dog.

We provide the binding through the form of personalization embeddings $f = \textrm{Concat}(f_1, \dots, f_n, \dots,  f_N)$, where $f_n$ encodes information from both the reference image and the corresponding entity word. Here, $N$ is the number of reference images. Specifically, to produce embeddings $f_n$, we first encode the image as image tokens $x_n \in \mathcal{R}^{l \times d} $, using a shared, frozen image encoder~\cite{dino_v2}. Here, $l$ denotes the number of tokens per reference image, and $d$ denotes the dimension of each token.

Next, we retrieve word tokens $c_n$ from the text embeddings $c$ (encoded from the text) and flatten $c_n$ into a 1-D embedding. Since the number of tokens of an entity word varies, we zero-pad or truncate the word embeddings to a consistent length. To bind the image and word tokens, we replicate the flattened word tokens $l$ times and concatenate them with the image tokens along the channel axis. Finally, we pass it to a linear projection module, apply a residual connection with the image tokens $x_n$, and add a learnable image index embedding to separate tokens from different images. Different tokens from the same image will share the same image index embedding.

\iftoggle{arxiv}{\vspace{2mm}}{}
\myparagraph{Personalization Conditioning.}
The personalization embeddings $f$ are then used to compute cross-attention with video tokens. Although IP-Adapter~\cite{ip_adapter} uses a single decoupled cross-attention layer for both text and image conditioning, we find empirically that separate cross-attention layers perform better in our case. This is likely because our multi-image conditioning introduces a longer sequence of image tokens. Thus, mixing text and image tokens in a shared layer causes the image tokens to dominate, reducing alignment with the text prompt.

We train the model in two stages. In the first stage, we train the model with only one cross-attention for text conditioning. Next, we introduce the additional cross-attention for personalization conditioning and fine-tune the whole model with warmup. Appendix~\ref{app:details_model} details model training.

\iftoggle{arxiv}{}{\vspace{-1.5mm}}
\subsection{Reducing Model Overfitting}
\label{sec:method_overfitting}
\iftoggle{arxiv}{}{\vspace{-1mm}}

We learn \model by denoising the training videos using the selected and segmented frames as conditions. However, this approach often leads to overfitting, where the model learns to focus on the lighting, pose, occlusion, and camera viewpoint of the reference subject (\textit{ref}) rather than its identity. Specifically, we find that:

\begin{itemize}[leftmargin=8pt]
    \item If \textit{ref} is high-resolution, the model generates a large subject close to the camera.
    \item If \textit{ref} is occluded, the model generates other objects that occlude the target subject.
    \item If \textit{ref} is cropped, the model places the subject at the edge, causing it to be cropped by the video boundary.
    \item The model often replicates the subject's pose and lighting conditions from \textit{ref}.
    \item If multiple \textit{ref}s represent the same subject with similar poses, the model produces a subject with minimal motion.
\end{itemize}

This overfitting often leads to the \textit{copy-and-paste} effect, where the model directly replicates the reference images in the video without introducing pose and lighting variations. This effect is commonly observed in reconstruction-based methods, such as IP-Adapter~\cite{ip_adapter}, as shown in Figure~\ref{fig:msrvtt_comparison}.

To alleviate these issues, we apply data augmentation to the reference images. Specifically, we use downscaling and Gaussian blurring to prevent overfitting to the image resolution, color jittering and brightness adjustment to mitigate overfitting on the lighting conditions, and random horizontal flip, image shearing, and rotation to weaken overfitting on the subject's pose. The key idea is to guide the model to focus on the subject's identity rather than learning the unintended information leakage from the reference images. More details on the proposed image augmentations can be found in Appendix~\ref{app:details_dataset_augmentations}.

%% file: figures/model_architecture.tex
\begin{figure*}[t]
    \centering
    \includegraphics[width=\linewidth]{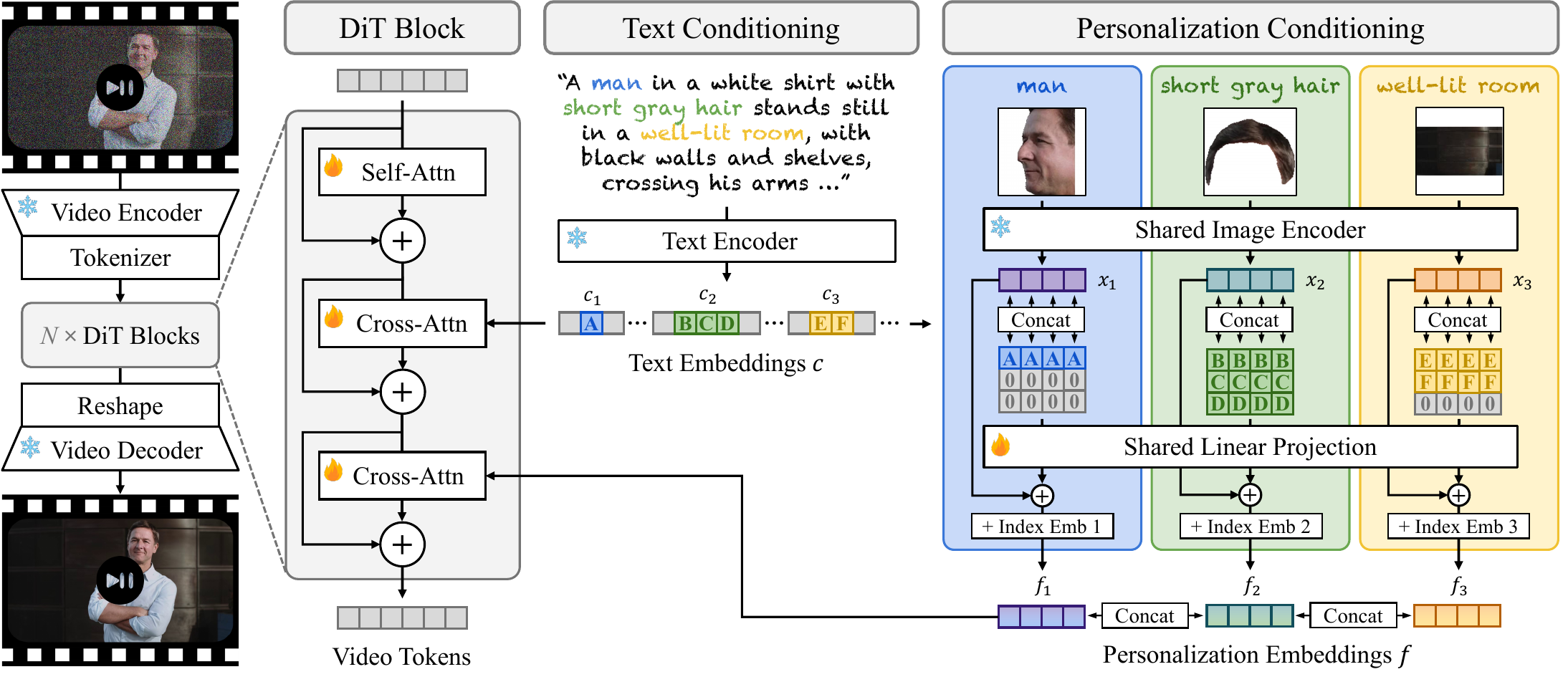}
    \iftoggle{arxiv}{\vspace{-6mm}}{\vspace{-8mm}}
    \mycaption{Model architecture}{
        Our model is a latent DiT~\cite{dit}, where we first encode a video into video tokens and denoise them with a deep cascade of DiT blocks in the latent space. Each DiT block includes an additional cross-attention operation with personalization embeddings $f = \textrm{Concat}(f_1, \dots, f_n, \dots,  f_N)$, where $f_n$ fuses the embeddings of both the reference image $x_n$ and the corresponding entity word $c_n$. Each square in the figure represents a 1-D token.
    }
    \label{fig:model_architecture}
    \iftoggle{arxiv}{}{\vspace{-5mm}}
\end{figure*}

%% file: sources/4_experiment.tex
\input{figures/msrvtt_sample}

\iftoggle{arxiv}{}{\vspace{-2mm}}
\section{Experiments}
\label{sec:experiment}
\iftoggle{arxiv}{}{\vspace{-1.5mm}}

Section~\ref{sec:exp_benchmark} introduces \benchmark, a comprehensive benchmark for personalization. Section~\ref{sec:exp_sota} provides quantitative and qualitative comparisons with state-of-the-art methods. Section~\ref{sec:exp_ablation} discusses the ablation study of our model training and architecture designs. Appendix~\ref{app:details_datasets} contains details of the training dataset and augmentations. Appendix~\ref{app:details_model} includes details of model architecture, training, and inference. Finally, we include more generated samples in Appendix~\ref{app:visualziation}.

\iftoggle{arxiv}{}{\vspace{-1.5mm}}
\subsection{MSRVTT-Personalization Benchmark}
\label{sec:exp_benchmark}
\iftoggle{arxiv}{}{\vspace{-1mm}}

Existing methods~\cite{dreambooth,ip_adapter,dreamvideo,storydiffusion} evaluate subject preservation using image similarity~\cite{arcface,clip,dino_v2} between reference and generated images or videos. However, these metrics are ineffective for multiple subjects, as image-level similarity fails to focus on the target subject. To address this issue, we introduce \benchmark to provide a more comprehensive and accurate evaluation of personalization tasks. It supports various conditioning scenarios, including conditioning on face crops, single or multiple subjects, and both foreground objects and backgrounds.

We construct the test benchmark based on MSR-VTT~\cite{msrvtt} and process the dataset in three steps. First, we use TransNetV2~\cite{transnetv2}, a shot boundary detection algorithm, to split long videos into multiple clips and apply an internal captioning algorithm to create detailed captions for each clip.  Next, we follow the procedure in Section~\ref{sec:method_dataset} to produce annotations for each video-caption pair. Finally, to ensure data quality, we manually select samples that meet the following criteria:

\begin{itemize}[leftmargin=8pt]
    \item Video is not an animated still image without meaningful subject motion.
    \item Video does not contain extensive text overlays.
    \item The retrieved subjects and objects cover all the main subjects and objects in the video.
    \item The background image, produced by an inpainting algorithm, has successfully removed foreground objects without generating new objects.
\end{itemize}

To increase data diversity, we select one clip from each long video and collect 2,130 clips. Figure~\ref{fig:msrvtt_sample} shows an annotated test sample.

\iftoggle{arxiv}{}{\input{tables/msrvtt_comparison}}

\iftoggle{arxiv}{\vspace{2mm}}{}
\myparagraph{Evaluation Metrics.}
An ideal personalized video output should align with the text, preserve subject fidelity, and exhibit natural video dynamics. Therefore, we use the following five metrics:

\begin{itemize}[leftmargin=8pt]
    \item Text similarity~\cite{godiva}: cosine similarity between the CLIP ViT-L/14~\cite{clip} features of the text and the generated frames. It measures how the generated video aligns with the text prompt. 
    \item Video similarity~\cite{textual_inversion}: average cosine similarity between the CLIP ViT-L/14 features of the ground truth and generated frames.
    \item Subject similarity: average cosine similarity between the DINO ViT-B/16~\cite{dino} features of the reference images and the segmented subject of the generated frames. We segment the subjects using Grounding-DINO Swin-T~\cite{grounding_dino} and SAM ViT-B/16~\cite{sam}.
    \item Face similarity: average cosine similarity between the ArcFace R100~\cite{arcface} features of the reference face crops and the generated face crops. We detect generated faces using YOLOv9-C~\cite{yolov9}.
    \item Dynamic degree~\cite{vbench}: optical flow magnitude between consecutive generated frames. We compute the optical flow using RAFT~\cite{raft}.
\end{itemize}

Note that video frames with missing subjects or faces are assigned a similarity score of $0$. The benchmark is publicly available at~\url{https://github.com/snap-research/MSRVTT-Personalization}.

\iftoggle{arxiv}{\input{tables/msrvtt_comparison}}{}
\input{figures/msrvtt_comparison}

\input{tables/user_study}
\iftoggle{arxiv}{\input{tables/ablation_study}}{}

\subsection{Comparisons with the State-of-the-Arts}
\label{sec:exp_sota}
\iftoggle{arxiv}{}{\vspace{-1.5mm}}

In this section, we quantitatively and qualitatively compare \model with existing personalization models on \benchmark.

\iftoggle{arxiv}{\vspace{2mm}}{}
\myparagraph{Experimental Setups.}
We extensively compare various personalization models, including text-to-image~\cite{elite,ip_adapter,photomaker} and text-to-video models~\cite{videobooth,dreamvideo,magicme}, as well as optimization-based~\cite{dreamvideo,magicme} and encoder-based methods~\cite{elite,videobooth,ip_adapter,photomaker}. As existing methods use different types of conditional images, we introduce two evaluation modes: subject mode and face mode. Subject mode uses full subject images as input, while face mode uses only face crops. For subject mode, we collect 1,736 test videos with a single subject. For face mode, we collect 1,285 test videos with a single person's face crop.

For text-to-image models~\cite{elite,ip_adapter,photomaker}, we treat the output images as single-frame videos. For optimization-based models~\cite{dreamvideo,magicme}, we use the default hyperparameters in the official codebase for finetuning. For IP-adapter~\cite{ip_adapter}, we use the checkpoint of IP-Adapter-FaceID+. If the model supports multiple reference images, we evaluate it with both single and multiple input images. We additionally evaluate our model with an extra input of a background reference image in the subject mode.

\iftoggle{arxiv}{}{\input{tables/ablation_study}}

\iftoggle{arxiv}{\vspace{2mm}}{}
\myparagraph{Quantitative Evaluation on \benchmark.}
Table~\ref{tab:msrvtt_comparison} shows the quantitative evaluation results. Compared to the existing open-set personalization methods~\cite{elite,videobooth,dreamvideo}, \model achieves higher subject fidelity, with a $23.2\%$ higher subject similarity than VideoBooth~\cite{videobooth}. Meanwhile, our model achieves the best text alignment and greatest video dynamics. Notably, our open-set model outperforms face-specific models~\cite{ip_adapter,photomaker,magicme} in face fidelity, achieving $11.3\%$ higher face similarity than IP-adapter~\cite{ip_adapter}.

Moreover, \model can generate the target subject or face with higher fidelity when provided with more reference images, demonstrating the advantage of multi-image conditioning. Furthermore, leveraging an extra background reference image, \model can synthesize a video more similar to the ground truth video, highlighting the effectiveness of our background conditioning. However, more reference images sometimes lead to worse textual alignment, potentially due to the limited flexibility introduced by more reference images.

\iftoggle{arxiv}{\vspace{4mm}}{}
\myparagraph{Qualitative Evaluation on \benchmark.}
In Figure~\ref{fig:msrvtt_comparison}, we show videos generated by different methods alongside the ground truth videos. More comparisons on various conditional subjects can be found in Appendix~\ref{app:visualziation_comparison}. Compared to existing models, our method produces more photorealistic videos with higher fidelity for target subjects.

\iftoggle{arxiv}{\vspace{4mm}}{}
\myparagraph{Human Evaluation.}
To complement automated evaluation, we conduct a user study to assess visual quality and subject fidelity. We randomly select 200 test samples from the subject and face modes, respectively, and show the conditional image and the results to $5$ participants. For each sample, participants are asked to select the one that best preserves the subject details and has the best visual quality. 

Table~\ref{tab:user_study} summarizes the results. Our method significantly outperforms the state-of-the-art methods in both visual quality and subject fidelity. Notably, the fidelity scores reported by humans are positively correlated to the scores of subject similarity and face similarity in Table~\ref{tab:msrvtt_comparison}, showcasing the effectiveness of the proposed \benchmark.

\iftoggle{arxiv}{}{\vspace{-1mm}}
\subsection{Ablation Study}
\label{sec:exp_ablation}
\iftoggle{arxiv}{}{\vspace{-0.5mm}}

In this section, we present an ablation study with three training or architecture choices. The quantitative and qualitative evaluations are shown in Table~\ref{tab:ablation_study} and Figure~\ref{fig:ablation_study}, respectively.

\iftoggle{arxiv}{\vspace{2mm}}{}
\myparagraph{Different Image Encoders.}
Inspired by Wu~\etal~\cite{neural_assets}, we train the models with two image encoders, CLIP~\cite{clip} and DINOv2~\cite{dino_v2}, and find that CLIP achieves better text similarity, while DINOv2 performs better in subject similarity. We hypothesize that DINOv2, trained with self-supervised learning objectives, captures unique object features. In contrast, CLIP, designed to bridge visual and textual modalities, focuses on details typically described in the prompt, which can improve the text-image alignment. 

\iftoggle{arxiv}{\vspace{7mm}}{}
\myparagraph{Necessity of Binding Image and Word Concepts.}
In Section~\ref{sec:method_framework}, we propose a mechanism to bind the concepts of images and the corresponding entity words. Without such binding, the model may incorrectly apply image conditions to the wrong subject. For example, the model places a reference human face on a dog as in the second row of Figure~\ref{fig:ablation_study}. This misalignment also results in missing subjects and lower subject similarity.

\iftoggle{arxiv}{\vspace{2mm}}{}
\myparagraph{Effect of Data Augmentation.}
In Section~\ref{sec:method_overfitting}, we introduce data augmentation to reduce model learning. Without augmentation, the model suffers from the \textit{copy-and-paste} issue. Although this helps to achieve higher subject similarity, it degrades dynamic degree and decreases text similarity. Specifically, although the prompt in Figure~\ref{fig:ablation_study} is \texttt{a woman is smiling ...}, the synthetic subject in the third row does not \textit{smile}. Instead, it replicates the same facial expression as in the reference image.

\input{figures/ablation_study}

%% file: figures/msrvtt_sample.tex
\begin{figure*}[t]
    \centering
    \includegraphics[width=\iftoggle{arxiv}{\linewidth}{.85\linewidth}]{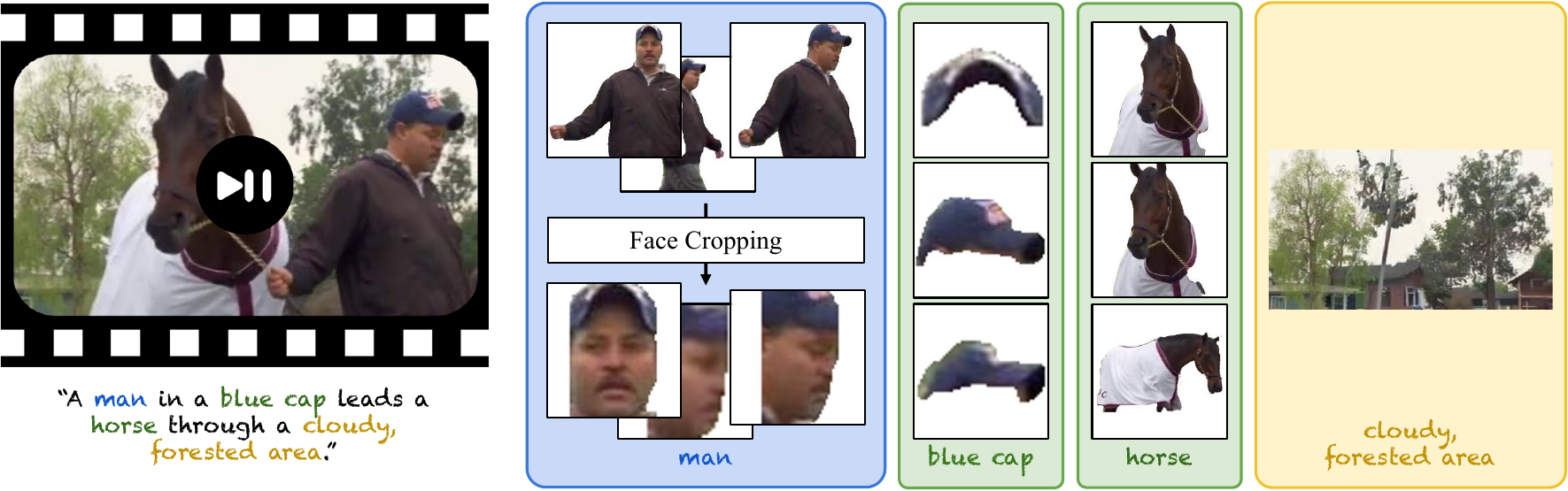}
    \iftoggle{arxiv}{\vspace{-4mm}}{\vspace{-2mm}}
    \mycaption{Test sample in \benchmark}{
         We present a comprehensive video personalization benchmark. Our benchmark supports various modes, including face conditioning, single or multiple subjects conditioning, and foreground and background conditioning.
    }
    \label{fig:msrvtt_sample}
    \iftoggle{arxiv}{}{\vspace{-2mm}}
\end{figure*}

%% file: tables/msrvtt_comparison.tex
\begin{table*}[t]
    \centering
    \iftoggle{arxiv}{\small}{\footnotesize}
    \iftoggle{arxiv}{}{\renewcommand{\arraystretch}{0.9}}
    \mycaption{Quantitative comparison on \benchmark}{
        We compare \model with state-of-the-art personalization methods across multiple metrics, including text similarity (Text-S), video similarity (Vid-S), subject similarity (Subj-S), face similarity (Face-S), and dynamic degree (Dync-D). The top and bottom tables show the evaluations for subject and face modes, respectively. $^\dagger$For text-to-image models, outputs are treated as single-frame videos without evaluating temporal quality. We evaluate \model with the videos at $512\mathrm{px} \times 288\mathrm{px}$ resolution. We highlight the top two models for the single reference image setting.
    }
    \label{tab:msrvtt_comparison}
    \iftoggle{arxiv}{\vspace{-1mm}}{\vspace{-2mm}}
    \begin{tabular}{lccccccc}
        \toprule
        \multirow{2}{*}{Method} &
        \multirow{2}{*}{\makecell{Test-time \\ Optimization}} &
        \multicolumn{2}{c}{Reference Images} &
        \multirow{2}{*}{Text-S$\uparrow$} &
        \multirow{2}{*}{Vid-S$\uparrow$} &
        \multirow{2}{*}{Subj-S$\uparrow$} &
        \multirow{2}{*}{Dync-D$\uparrow$} \\ \cmidrule(lr){3-4}
                                     &        & Subject  & Background & & & \\
        \midrule \midrule
        ELITE$^\dagger$~\cite{elite} & \xmark & single   & \xmark & 0.245 & \cellsecond 0.620 & 0.359 & - \\
        VideoBooth~\cite{videobooth} & \xmark & single   & \xmark & 0.222 & 0.612 & \cellsecond 0.395 & \cellsecond 0.448 \\
        DreamVideo~\cite{dreamvideo} & \cmark & single   & \xmark & \cellsecond 0.261 & 0.611 & 0.310 & 0.311 \\
        \model                       & \xmark & single   & \xmark & \cellfirst 0.269 & \cellfirst 0.732 & \cellfirst 0.617 & \cellfirst 0.466 \\
        \midrule
        DreamVideo~\cite{dreamvideo} & \cmark & multiple & \xmark & 0.253 & 0.604 & 0.256 & 0.303 \\
        \model                       & \xmark & multiple & \xmark & 0.268 & 0.743 & 0.626 & 0.473 \\
        \midrule
        \model                       & \xmark & multiple & \cmark & 0.254 & 0.780 & 0.570 & 0.506 \\
        \bottomrule
    \end{tabular}
    \\[+1mm]
    \begin{tabular}{lcccccc}
        \toprule
        \multirow{2}{*}{Method} &
        \multirow{2}{*}{\makecell{Test-time \\ Optimization}} &
        Reference Images &
         \multirow{2}{*}{Text-S$\uparrow$} &
        \multirow{2}{*}{Vid-S$\uparrow$} &
        \multirow{2}{*}{Face-S$\uparrow$} &
        \multirow{2}{*}{Dync-D$\uparrow$} \\ \cmidrule(lr){3-3}
                                               &        & Face Crop & & & \\
        \midrule \midrule
        IP-Adapter$^\dagger$~\cite{ip_adapter} & \xmark & single    & 0.251 & \cellsecond 0.648 & \cellsecond 0.269 & - \\
        PhotoMaker$^\dagger$~\cite{photomaker} & \xmark & single    & \cellfirst 0.278 & 0.569 & 0.189 & - \\
        Magic-Me~\cite{magicme}                & \cmark & single    & 0.251 & 0.602 & 0.135 & \cellsecond 0.418 \\
        \model                                 & \xmark & single    & \cellsecond 0.273 & \cellfirst 0.687 & \cellfirst 0.382 & \cellfirst 0.424 \\
        \midrule
        PhotoMaker$^\dagger$~\cite{photomaker} & \xmark & multiple  & 0.275 & 0.582 & 0.216 & -     \\
        Magic-Me~\cite{magicme}                & \cmark & multiple  & 0.248 & 0.618 & 0.153 & 0.385 \\
        \model                                 & \xmark & multiple  & 0.272 & 0.694 & 0.411 & 0.402 \\
        \bottomrule
    \end{tabular}
    \iftoggle{arxiv}{\vspace{-1mm}}{\vspace{-3mm}}
\end{table*}

%% file: figures/msrvtt_comparison.tex
\begin{figure*}[t]
    \centering
    \includegraphics[trim={3mm 0 0 0}, width=\linewidth]{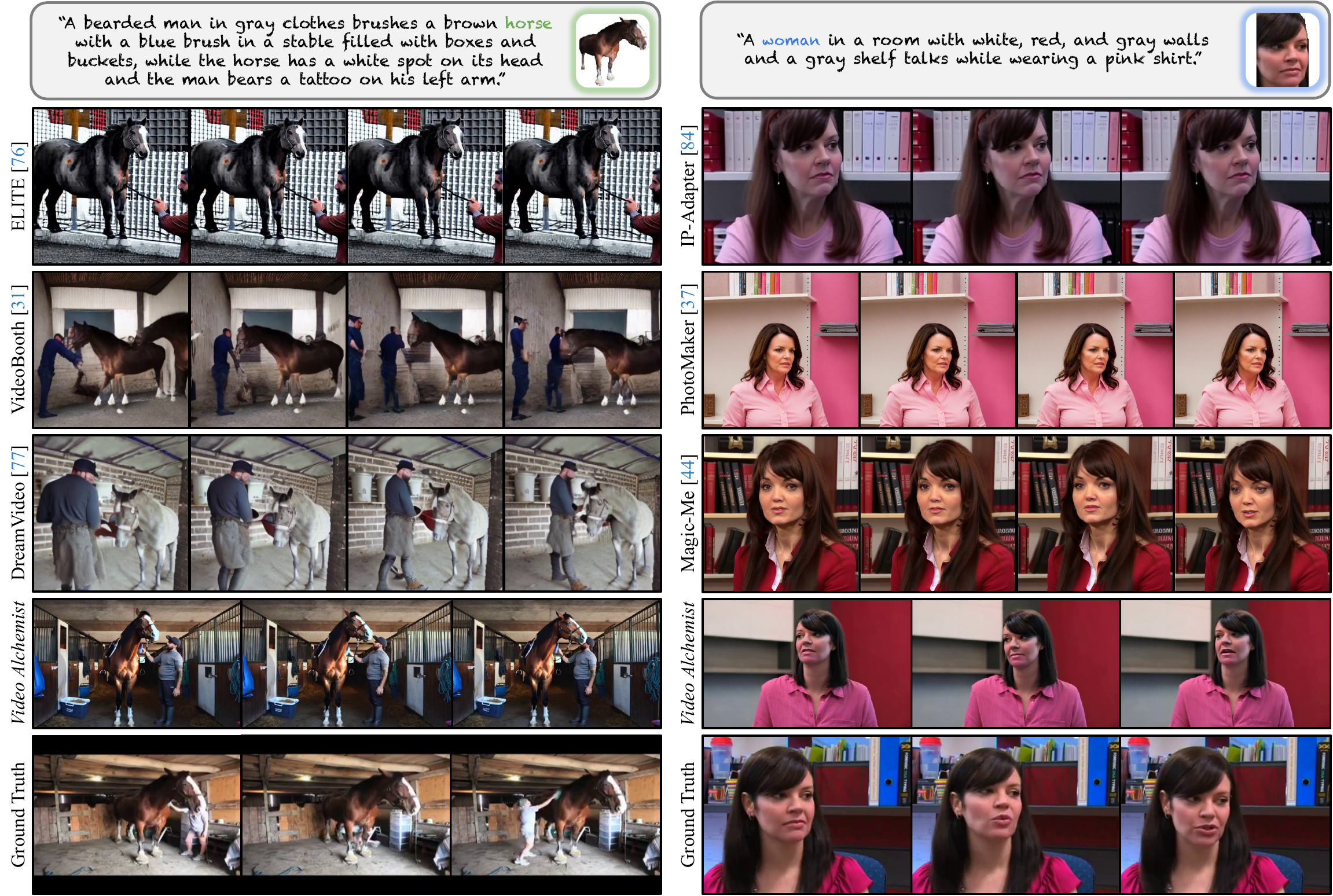}
    \iftoggle{arxiv}{\vspace{-4.5mm}}{\vspace{-6.5mm}}
    \mycaption{Qualitative comparison on \benchmark}{
        We use a single reference image to each model for a fair comparison. Compared to existing methods, our results closely match the input text prompt and reference subjects while exhibiting natural motion and pose variations.
    }
    \iftoggle{arxiv}{}{\vspace{-5mm}}
    \label{fig:msrvtt_comparison}
\end{figure*}

%% file: tables/user_study.tex
\begin{table}[t]
    \centering
    \footnotesize
    \setlength\tabcolsep{0.5pt}
    \mycaption{User preference study}{
        We show the user preference percentage for subject (left) and face modes (right), respectively.
    }
    \label{tab:user_study}
    \iftoggle{arxiv}{\vspace{-1mm}}{\vspace{-2mm}}
    \begin{minipage}[t]{.495\linewidth}
        \vspace{0pt}
        \begin{tabular}{lcc}
            \toprule
            \multirow{2}{*}{Method} &
            \multicolumn{2}{c}{Preference Ratio$\uparrow$} \\ \cmidrule(lr){2-3}
                                         & Quality & Fidelity \\
            \midrule
            ELITE~\cite{elite}           & 2.7\% \cellsecond & 0.6\% \\
            VideoBooth~\cite{videobooth} & 0.3\% & 0.8\% \cellsecond \\
            DreamVideo~\cite{dreamvideo} & 0.5\% & 0.5\% \\
            \model                       & 96.5\% \cellfirst & 98.1\% \cellfirst \\
            \bottomrule
        \end{tabular}
    \end{minipage}
    \hfill
    \begin{minipage}[t]{.495\linewidth}
        \vspace{0pt}
        \begin{tabular}{lcc}
            \toprule
            \multirow{2}{*}{Method} &
            \multicolumn{2}{c}{Preference Ratio$\uparrow$} \\ \cmidrule(lr){2-3}
                                         & Quality & Fidelity \\
            \midrule
            IP-Adapter~\cite{ip_adapter} & 10.4\% & \cellsecond 20.2\% \\
            PhotoMaker~\cite{photomaker} & \cellsecond 37.5\% & 7.4\% \\
            Magic-Me~\cite{magicme}      & 4.4\% & 4.0\% \\
            \model                       & \cellfirst 47.6\% & \cellfirst 68.4\% \\
            \bottomrule
        \end{tabular}
    \end{minipage}
    \iftoggle{arxiv}{\vspace{-1mm}}{\vspace{-5mm}}
\end{table}

%% file: tables/ablation_study.tex
\begin{table*}[t]
    \centering
    \iftoggle{arxiv}{\small}{\footnotesize}
    \mycaption{Ablation study for the subject mode}{
        We use a single reference image for each model and examine three control factors. The experiments are conducted on the videos at $256\mathrm{px} \times 144\mathrm{px}$ resolution.
    }
    \label{tab:ablation_study}
    \iftoggle{arxiv}{\vspace{-1mm}}{\vspace{-3mm}}
    \begin{tabular}{lcccccccc}
        \toprule
        Method          & Image Encoder & Use Word Token & Image Augmentations & Text-S$\uparrow$ & Vid-S$\uparrow$ & Subj-S$\uparrow$ & Dync-D$\uparrow$ \\
        \midrule \midrule
        Use CLIP        & CLIP~\cite{clip}      & \cmark & \cmark & \cellfirst 0.269 & 0.768 & 0.569 & 0.552 \\
        No word token   & DINOv2~\cite{dino_v2} & \xmark & \cmark & 0.256 & \cellfirst 0.790 & 0.566 & \cellsecond 0.569 \\
        No augmentation & DINOv2~\cite{dino_v2} & \cmark & \xmark & 0.251 & \cellsecond 0.781 & \cellfirst 0.609 & 0.506 \\
        \model          & DINOv2~\cite{dino_v2} & \cmark & \cmark & \cellsecond 0.257 & \cellfirst 0.790 & \cellsecond 0.600 & \cellfirst 0.570 \\
        \bottomrule
    \end{tabular}
    \iftoggle{arxiv}{\vspace{-1mm}}{\vspace{-2mm}}
\end{table*}

%% file: figures/ablation_study.tex
\begin{figure}[t]
    \centering
    \includegraphics[trim={4mm 0 0 0}, width=\linewidth]{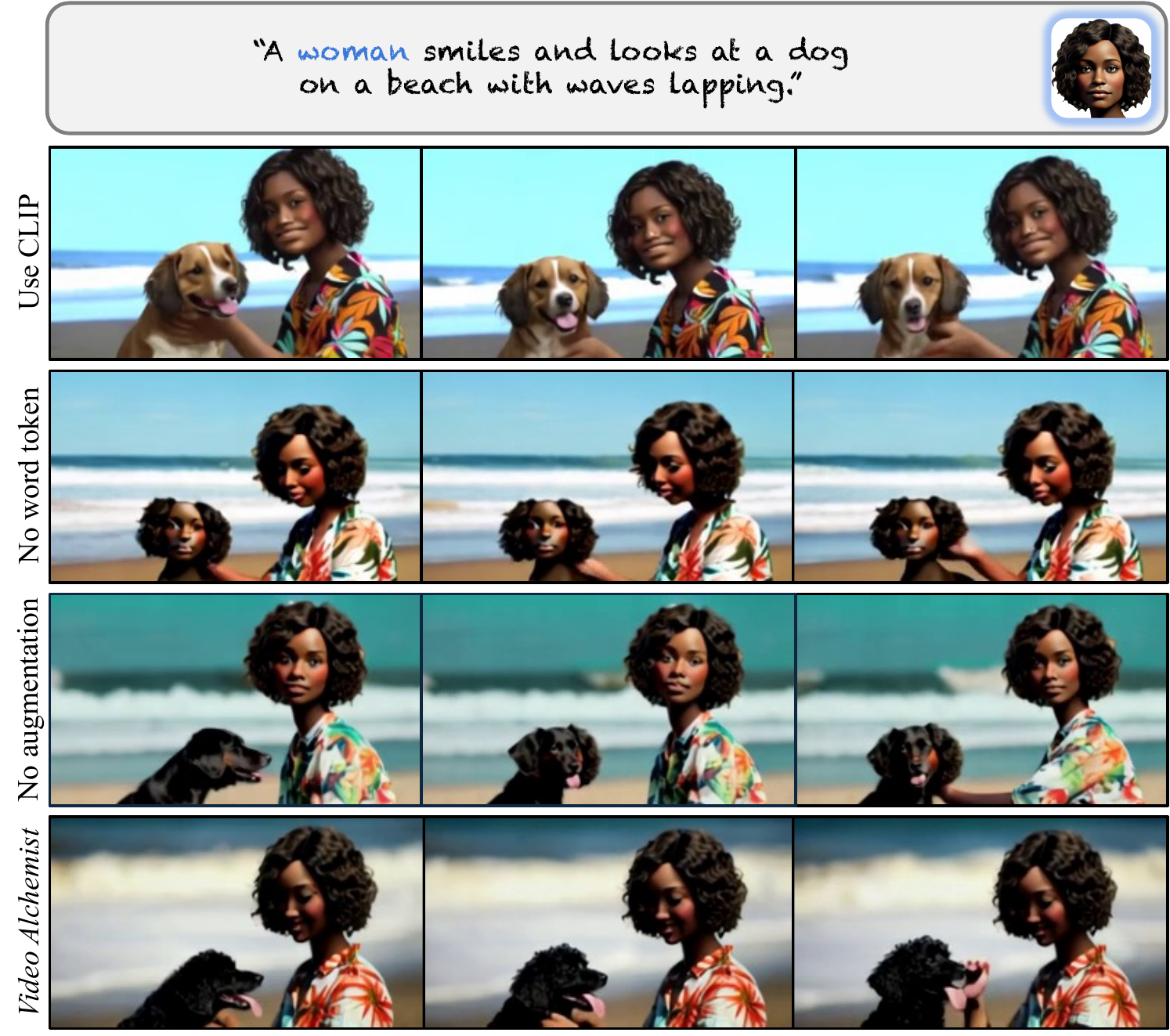}
    \iftoggle{arxiv}{\vspace{-4mm}}{\vspace{-7mm}}
    \mycaption{Qualitative results of the ablation study}{
        From top to bottom, we show that 1) \model achieves better subject fidelity using DINOv2~\cite{dino_v2} as the image encoder; 2) it correctly binds the conditional image and entity word with the usage of word tokens; 3) it mitigates the \textit{copy-and-paste} effect and synthesizes text-aligned videos via the proposed data augmentation. The reference image is synthesized by DALL·E 3~\cite{dalle3}.
    }
    \label{fig:ablation_study}
    \iftoggle{arxiv}{}{\vspace{-3mm}}
\end{figure}

%% file: sources/5_conclusion.tex
\iftoggle{arxiv}{}{\vspace{-1.5mm}}
\section{Conclusion}
\label{sec:conclusion}
\iftoggle{arxiv}{}{\vspace{-1mm}}

We have presented \model, a video personalization model that supports multi-subject and open-set personalization capabilities for both foreground objects and background without requiring test-time optimization. It is built on a Diffusion Transformer module that integrates conditional images with their subject-level prompts through cross-attention layers. With our dataset curation and data augmentation, we have reduced model overfitting on undesirable properties of the reference images. In addition, we have introduced a new benchmark for evaluating personalization models across various conditioning scenarios. Experimental results show that our method outperforms existing methods in quantitative and qualitative measures.

\iftoggle{arxiv}{\vspace{2mm}}{\clearpage}
\myparagraph{Acknowledgments.} Supported by the Intelligence Advanced Research Projects Activity (IARPA) via Department of Interior/ Interior Business Center (DOI/IBC) contract number 140D0423C0074. The U.S. Government is authorized to reproduce and distribute reprints for Governmental purposes notwithstanding any copyright annotation thereon. Disclaimer: The views and conclusions contained herein are those of the authors and should not be interpreted as necessarily representing the official policies or endorsements, either expressed or implied, of IARPA, DOI/IBC, or the U.S. Government.

We thank Ziyi Wu, Moayed Haji Ali, and Alper Canberk for their helpful discussions.

%% file: sources/A_detail_dataset.tex
\section{Details of Training Datasets and Augmentations}
\label{app:details_datasets}

\subsection{Training Datasets and Undesirable Samples Filtering}
Our personalization training dataset is built on Panda-70M~\cite{panda} and other internal video-caption datasets, consisting of 86.8M videos. However, the original dataset includes undesirable video samples for video generation. We classify these undesirable samples into four categories:

\begin{itemize}[leftmargin=8pt]
    \item Still foreground image: a video with only pan and zoom effects of a static image.
    \item Slight motion: a video with tiny camera movement and static foreground objects
    \item Screen-in-screen: a video with an image or video overlaying on a background image or video.
    \item Computer screen recording: a video depicting a screen recording (excluding PC games).
\end{itemize}

To filter out these samples, we learn a video classification model. Specifically, we randomly sample 40k videos from our training dataset and manually annotate them based on the above criteria. Using these labels, we fine-tune VideoMAE~\cite{videomae} for video classification.  Moreover, as we aim to generate single-shot videos, we apply TransNetV2~\cite{transnetv2} to detect and exclude videos that contain multiple shots. We only retain the desirable single-shot videos for training.

\subsection{Retrieving Entity Words}
\label{app:details_datasets_retrieval}
In Section~\ref{sec:method_dataset}, we use a large language model~\cite{mistral} (LLM) to retrieve the entity words from the caption, using the instruction template shown in Figure~\ref{fig:prompt}.

\input{figures/prompt}

Given the caption, the LLM extracts a list of entity words, with the following steps.

\begin{itemize}[leftmargin=8pt]
    \item Remove the sample if the output of the LLM is not in a valid dictionary format.
    \item Remove the sample if any entity word is not a sub-string of the caption.
    \item Reclassify the entity words according to the pre-defined rules. For example, ``cloud'' is not a visually separable object and is supposed to be classified into a background entity word.
    \item Remove the sample with no subject entity word, as we observe that the video motion of these samples is typically trivial camera movements and lacks meaningful foreground motion.
    \item Remove the sample with the subject entity word in the plural form, as this will introduce ambiguity when applying the localization algorithm.
\end{itemize}

We curate a training dataset comprising 37.8M videos. To illustrate the diversity of subjects within our dataset, we plot a word cloud of entity words from 10k randomly sampled training videos in Figure~\ref{fig:word_cloud}.

\input{figures/word_cloud}

\subsection{Data Augmentation and Conditional Images Sampling}
\label{app:details_dataset_augmentations}
In Section~\ref{sec:method_overfitting}, we introduce data augmentation to prevent models from overly relying on the undesirable properties of the reference image. Table~\ref{tab:augmentation} lists the applied augmentation. While augmentations can reduce model overfitting to some extent, we observe that models could also overfit to the number of reference images. Specifically, if we always use all available reference images as conditions during training, the model can generate the target subject with some properties correlated to the number of reference images (\textit{ref}) during inference. Using the text prompt ``\textit{A dog is running}'' as an example:

\begin{itemize}[leftmargin=8pt]
    \item If users input 0 \textit{ref}, the model generates a tiny or heavily occluded dog.
    \item If users input 1 \textit{ref}, the model generates a dog that is running out of view of the video.
    \item If users input 3 \textit{ref}s of a similar pose, the model generates a dog that is running in slow motion.
\end{itemize}

To avoid models overfitting on the number of the reference images, we design a sampling algorithm to select conditional subjects and their reference images during training. It includes the following five steps:

\begin{itemize}[leftmargin=8pt]
    \item Randomly sample the number of conditional subjects from 1 to 3.
    \item Randomly sample conditional subjects with replacement.
    \item For each subject, randomly sample the number of conditional reference images from 1 to 3.
    \item For each subject, randomly sample conditional reference images with replacement.
    \item Randomly include background conditioning with a probability of $50\%$.
\end{itemize}

\input{tables/augmentation}

%% file: figures/prompt.tex
\begin{figure}[h]
    \centering
    \includegraphics[width=\linewidth]{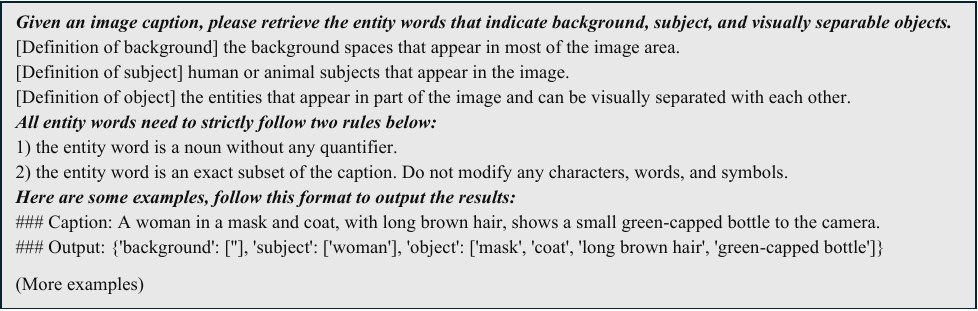}
    \mycaption{Prompt template for retrieving the entity words}{}
    \label{fig:prompt}
\end{figure}

%% file: figures/word_cloud.tex
\begin{figure}[t]
    \centering
    \includegraphics[width=\linewidth]{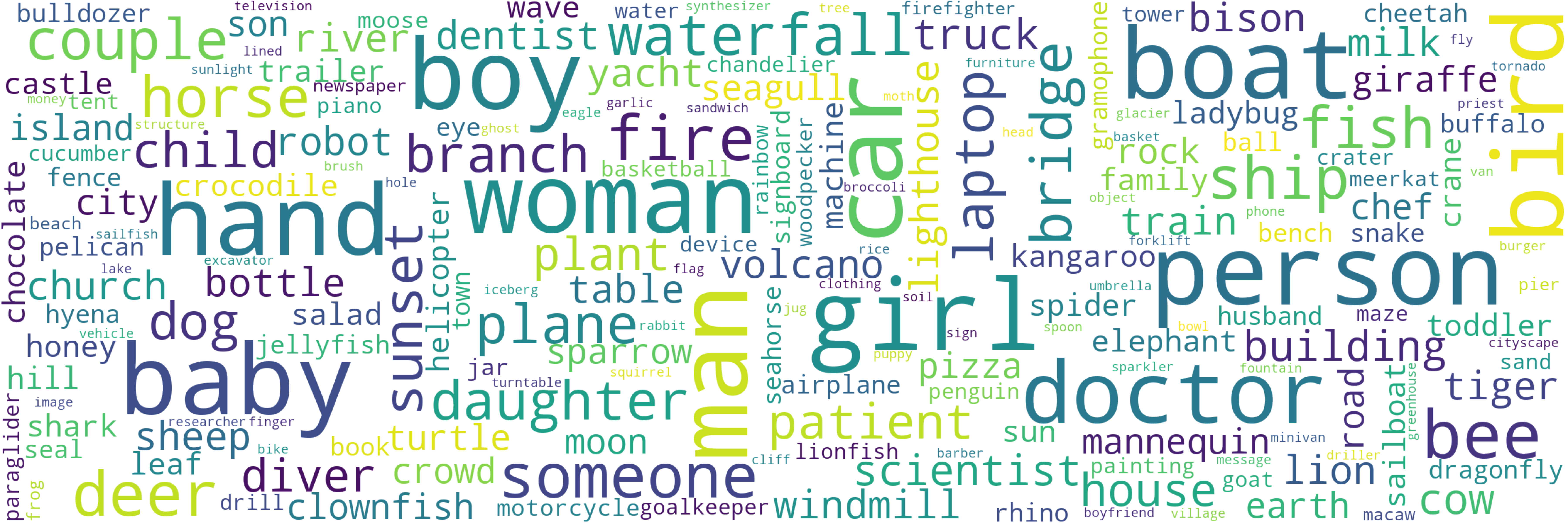}
    \vspace{-3mm}
    \mycaption{Word cloud of the entity words}{
        We randomly sample 10k videos from our training dataset and plot the word cloud of the retrieved subject and object entity words.
    }
    \label{fig:word_cloud}
\end{figure}

%% file: tables/augmentation.tex
\begin{table}[t]
    \centering
    \mycaption{Training augmentations}{
        We denote the height and width of the reference image as $h$ and $w$.
    }
    \label{tab:augmentation}
    \begin{tabular}{lccc}
        \toprule
        &
        \multirow{2}{*}{\makecell{Apply \\ Probability}} &
        \multicolumn{2}{c}{Hyperparameters} \\ \cmidrule(lr){3-4}
                          &       & Type                        & Sampling Range           \\
        \midrule
        Downscale         & $1.0$ & scale                       & $[112/\max(h, w), 1.0]$  \\
        Gaussian blur     & $1.0$ & kernel size ($\mathrm{px}$) & $[1, \max(h,w)/50]$      \\
        Color jitter      & $1.0$ & scale                       & $[-0.05, 0.05]$          \\
        Brightness        & $1.0$ & scale                       & $[0.9, 1.1]$             \\
        Horizontal flip   & $0.5$ & -                           & -                        \\
        Shearing (x-axis) & $1.0$ & value ($\mathrm{px}$)       & $[-0.05, 0.05] \times w$ \\
        Shearing (y-axis) & $1.0$ & value ($\mathrm{px}$)       & $[-0.05, 0.05] \times h$ \\
        Rotation          & $1.0$ & value ($^\circ$)            & $[-20, 20]$              \\
        Random crop       & $1.0$ & scale                       & $[0.67, 1.0]$            \\
        \bottomrule
    \end{tabular}
\end{table}

%% file: sources/B_detail_model.tex
\section{Details of Model Architecture, Training, and Inference}
\label{app:details_model}

\input{tables/architecture}
\input{tables/training}

\subsection{Model Architecture}
\label{app:details_model_architecture}
Our framework is a latent-based diffusion model. We use CogVideoX-5B~\cite{cogvideox} as the autoencoder with a compression rate of $4 \times 8 \times 8$ in temporal, height, and width dimensions. We use DiT~\cite{dit} as the video backbone with two different image encoders, including CLIP~\cite{clip} and DINOv2~\cite{dino_v2}. We detail the hyperparameters of the video backbone and image encoders in Table~\ref{tab:architecture}. For the video backbone, we follow the original DiT designs to embed input timesteps using adaLN-Zero block, which is composed of adaptive normalization layers~\cite{adaln} with scaling parameters $\alpha$ that are applied immediately prior to any residual connections within the DiT block. For the image representations, we find that using the patch tokens as the image embeddings can retain more localized properties of the reference images and result in higher fidelity than the class token.

\subsection{Model Training}
\label{app:details_model_training}
We present the training details of the model in Table~\ref{tab:training}. We train the model in two stages. In the first stage, we fix the autoencoder and train the video backbone without the cross-attention layer for personalization for 60k steps. In the second stage, we introduce the cross-attention layer for personalization and fine-tune the model for additional 40k steps. With more details, in the second stage, we apply a 1k-step linear warmup and only train the newly introduced cross-attention layer while keeping the video backbone fixed at the first 10k steps. For the following 30k steps, we fine-tune the entire video model with the image encoder frozen. We use the AdamW~\cite{adamw} optimizer with a constant learning rate of $1e^{-4}$. To achieve stable training, we set $\beta = [0.9, 0.99]$, a weight decay of $0.01$, gradient clipping with the value of $0.05$. We randomly drop text or image conditioning with a probability of $10\%$ and set them to zero to support classifier-free guidance~\cite{cfg}.

To enable the generation of high-resolution and long-duration videos while ensuring efficient model training, we train our model on videos of varying resolutions and lengths. Table~\ref{tab:training} lists the batch size and sampling weights for the training videos across different resolutions and lengths. The batch size is set to balance the training time for each step with different attributes. We apply the fixed framerate of $24$. Our model supports generating videos up to 12 seconds in length at $256\mathrm{px} \times 144\mathrm{px}$ resolution and up to 5 seconds in length at $512\mathrm{px} \times 288\mathrm{px}$ and $1024\mathrm{px} \times 576\mathrm{px}$ resolution.

Our model is implemented in PyTorch~\cite{pytorch} and trained with 256 80GB A100 GPUs in stage I and 64 GPUs in stage II.

\subsection{Model Inference}
\label{app:details_model_inference}
We use a rectified flow sampler~\cite{rfflow} with classifier-free guidance~\cite{cfg} (CFG) for sampling. The choice of scale and implementation of the CFG can significantly impact the performance of diffusion models. Although our model performs best with a CFG scale of $8$ for text conditioning, we find that applying such a large CFG scale for image conditioning can cause the model to replicate reference images directly into the video, without introducing natural motion and appearance variation. To address this, we follow Brooks~\etal~\cite{instructpix2pix} and apply CFG twice within a sampling step, once for text conditioning and once for image conditioning, but with a slight change in CFG implementation. Formally, Brooks~\etal~\cite{instructpix2pix} applies CFG as follows:

\vspace{-2.5mm}
\begin{equation}
    \tilde{e_{\theta}}(z_t, c_I, c_T) = e_{\theta}(z_t, c_I, c_T) + s_T \cdot (e_{\theta}(z_t, c_I, c_T) - e_{\theta}(z_t, c_I, \varnothing)) + s_I \cdot (e_{\theta}(z_t, c_I, \varnothing) - e_{\theta}(z_t, \varnothing, \varnothing)),
\end{equation}
where $e_{\theta}(z_t, c_I, c_T)$ is the score estimation function with the image and text conditioning, denoted as $c_I$ and $c_T$. We mark $c=\varnothing$ if we set condition $c$ to zero. Empirically, we find that the formula below can achieve better visual quality in our case:

\vspace{-2.5mm}
\begin{equation}
    \tilde{e_{\theta}}(z_t, c_I, c_T) = e_{\theta}(z_t, c_I, c_T) + s_T \cdot (e_{\theta}(z_t, c_I, c_T) - e_{\theta}(z_t, c_I, \varnothing)) + s_I \cdot (e_{\theta}(z_t, c_I, c_T) - e_{\theta}(z_t, \varnothing, c_T)).
\end{equation}

We set $s_T = 8$ and $s_I = 3$. We use $256$, $128$, and $64$ denoising steps to synthesize the videos at $256\mathrm{px} \times 144\mathrm{px}$, $512\mathrm{px} \times 288\mathrm{px}$, and $1024\mathrm{px} \times 576\mathrm{px}$ resolution, respectively. Moreover, we apply time shifting~\cite{scaling_rfflow,lumina} to align the signal-to-noise ratio (SNR) at different resolutions.

%% file: tables/architecture.tex
\begin{table}[t]
    \centering
    \mycaption{Architecture details of video generation backbone and image encoders}{}
    \label{tab:architecture}
    \begin{minipage}[t]{0.45\textwidth}
        \centering
        \vspace{+.1mm}
        \begin{tabular}{lc}
            \toprule
            Video Backbone          & DiT~\cite{dit}             \\
            \midrule
            Input channels          & 16                         \\
            Patch size              & $1 \times 2 \times 2$      \\
            Latent token channels   & 4096                       \\
            Positional embeddings   & RoPE                       \\
            DiT blocks count        & 32                         \\
            Attention heads count   & 32                         \\
            Use flash attention     & \cmark                     \\
            Use fused layer norm    & \cmark                     \\
            Use self conditioning   & \cmark                     \\
            Conditioning channels   & 1024                       \\
            Conditioning images     & 6 (stage II training only) \\
            \bottomrule
        \end{tabular}
    \end{minipage}
    \begin{minipage}[t]{0.4\textwidth}
        \centering
        \vspace{+.1mm}
        \begin{tabular}{lccc}
            \toprule
            Image Encoder    & CLIP~\cite{clip} & DINOv2~\cite{dino_v2} \\
            \midrule
            Architecture     & ViT-L/14          & ViT-L/14             \\
            Selective block  & 23                & 24                   \\
            Selective tokens & patch             & patch                \\
            Tokens count     & 256               & 256                  \\
            Tokens channels  & 1024              & 1024                 \\
            \bottomrule
        \end{tabular}
    \end{minipage}
    \vspace{+2mm}
\end{table}

%% file: tables/training.tex
\begin{table}[t!]
    \centering
    \mycaption{Training hyperparameters}{
        The right table is for stage II training.
    }
    \label{tab:training}
    \begin{minipage}[t]{0.4\textwidth}
        \centering
        \vspace{+.1mm}
        \begin{tabular}{lccc}
            \toprule
            Stage                & I      & II &                     \\
            \midrule
            Steps                & 60k    & 40k                      \\
            Warmup steps         & -      & 1k                       \\
            Samples seen         & 234M   & 39M                      \\
            Image conditioning   & \xmark & \cmark                   \\ \midrule
            Optimizer            & \multicolumn{3}{c}{AdamW}         \\
            Learning rate        & \multicolumn{3}{c}{$1e^{-4}$}     \\
            LR scheduler         & \multicolumn{3}{c}{constant}      \\
            Beta                 & \multicolumn{3}{c}{$[0.9, 0.99]$} \\
            Weight decay         & \multicolumn{3}{c}{$0.01$}        \\
            Gradient clipping    & \multicolumn{3}{c}{$0.05$}        \\
            Dropout              & \multicolumn{3}{c}{$0.1$}         \\
            \bottomrule
        \end{tabular}
    \end{minipage}
    \begin{minipage}[t]{0.55\textwidth}
        \centering
        \setlength\tabcolsep{3pt}
        \vspace{+.1mm}
        \begin{tabular}{crlrlrl}
            \toprule
            \# frames & \multicolumn{6}{c}{Batch Size (Sampling Weight)} \\ \midrule
            & \multicolumn{2}{c}{$256\mathrm{px} \times 144\mathrm{px}$}
            & \multicolumn{2}{c}{$512\mathrm{px} \times 288\mathrm{px}$}
            & \multicolumn{2}{c}{$1024\mathrm{px} \times 576\mathrm{px}$} \\
            $17$  & \hspace{+2mm} 1,216 & \hspace{-2mm} (10\%)  & \hspace{+2mm} 304 & \hspace{-2mm} (10\%)   & \hspace{+5mm} 80 & \hspace{-2mm} (10\%)  \\
            $49$  & \hspace{+2mm} 464   & \hspace{-2mm} (3.3\%) & \hspace{+2mm} 112 & \hspace{-2mm} (5.8\%)  & \hspace{+5mm} 32 & \hspace{-2mm} (5.8\%) \\
            $73$  & \hspace{+2mm} 320   & \hspace{-2mm} (3.3\%) & \hspace{+2mm} 80  & \hspace{-2mm} (5.8\%)  & \hspace{+5mm} 16 & \hspace{-2mm} (5.8\%) \\
            $97$  & \hspace{+2mm} 240   & \hspace{-2mm} (3.3\%) & \hspace{+2mm} 64  & \hspace{-2mm} (5.8\%)  & \hspace{+5mm} 16 & \hspace{-2mm} (5.8\%) \\
            $121$ & \hspace{+2mm} 192   & \hspace{-2mm} (3.3\%) & \hspace{+2mm} 48  & \hspace{-2mm} (5.8\%)  & \hspace{+5mm} 16 & \hspace{-2mm} (5.8\%) \\
            $145$ & \hspace{+2mm} 160   & \hspace{-2mm} (3.3\%) & -                 & \hspace{-2mm} (0\%)    & \hspace{+5mm} -  & \hspace{-2mm} (0\%)   \\
            $193$ & \hspace{+2mm} 128   & \hspace{-2mm} (3.3\%) & -                 & \hspace{-2mm} (0\%)    & \hspace{+5mm} -  & \hspace{-2mm} (0\%)   \\
            $289$ & \hspace{+2mm} 80    & \hspace{-2mm} (3.3\%) & -                 & \hspace{-2mm} (0\%)    & \hspace{+5mm} -  & \hspace{-2mm} (0\%)   \\
            \bottomrule
        \end{tabular}
    \end{minipage}
\end{table}

%% file: sources/C_visualization.tex
\section{More Visualization Results}
\label{app:visualziation}

In this section, we provide more synthetic samples to complement the evaluations. Appendix~\ref{app:visualziation_multi_subject} shows the samples of multi-subject and open-set customization. Appendix~\ref{app:visualziation_ablation} includes an ablation study in which we use different reference images to personalize the same conditional entity word from the same prompt. Appendix~\ref{app:visualziation_comparison} provides more comparisons with state-of-the-art personalization models on various conditional subjects.

\subsection{Additional Results of Multi-subject Open-set Personalization}
\label{app:visualziation_multi_subject}
We show the multi-subject and open-set personalization samples in Figures~\ref{fig:multi_subject_dinosaur} to~\ref{fig:multi_subject_rocket}. In each sample, we show the generated videos with one to three conditional subjects or backgrounds by incrementally increasing the number of reference images. In addition, we provide synthetic videos without reference images at the bottom to showcase the effect of image conditioning.

\clearpage
\input{figures/multi_subject_dinosaur}
\input{figures/multi_subject_discuss}
\input{figures/multi_subject_drink_tea}
\input{figures/multi_subject_rocket}

\clearpage
\subsection{Same Text Prompt with Different Reference Images}
\label{app:visualziation_ablation}
Figure~\ref{fig:teaser} presents videos generated using the same prompt and conditional subjects but varying background reference images. To demonstrate our model's robustness and ability to generate diverse visual content and motion, we showcase generated videos where the reference image for one subject is altered while keeping all other conditional inputs unchanged. Specifically, we provide samples with different reference images of \textit{person} in Figure~\ref{fig:ablate_person} and \textit{dog} in Figure~\ref{fig:ablate_dog}.

\input{figures/ablate_person}
\input{figures/ablate_dog}

\clearpage
\subsection{More Comparisons on Different Conditional Subjects}
\label{app:visualziation_comparison}
Figure~\ref{fig:msrvtt_comparison} shows qualitative comparisons between \model and state-of-the-art personalization models on the conditional subjects of \textit{horse} and \textit{woman}. In this section, we present more qualitative comparisons on other conditional subjects, including \textit{dog} in Figure~\ref{fig:subject_mode_dog}, \textit{cat} in Figure~\ref{fig:subject_mode_cat}, \textit{car} in Figure~\ref{fig:subject_mode_car}, and \textit{dinosaur toy} in Figure~\ref{fig:subject_mode_dinosaur}.

\input{figures/subject_mode_dog}
\input{figures/subject_mode_cat}
\input{figures/subject_mode_car}
\input{figures/subject_mode_dinosaur}

%% file: figures/multi_subject_dinosaur.tex
\begin{figure*}[t]
    \centering
    \includegraphics[trim={0 7mm 0 0}, width=.88\linewidth]{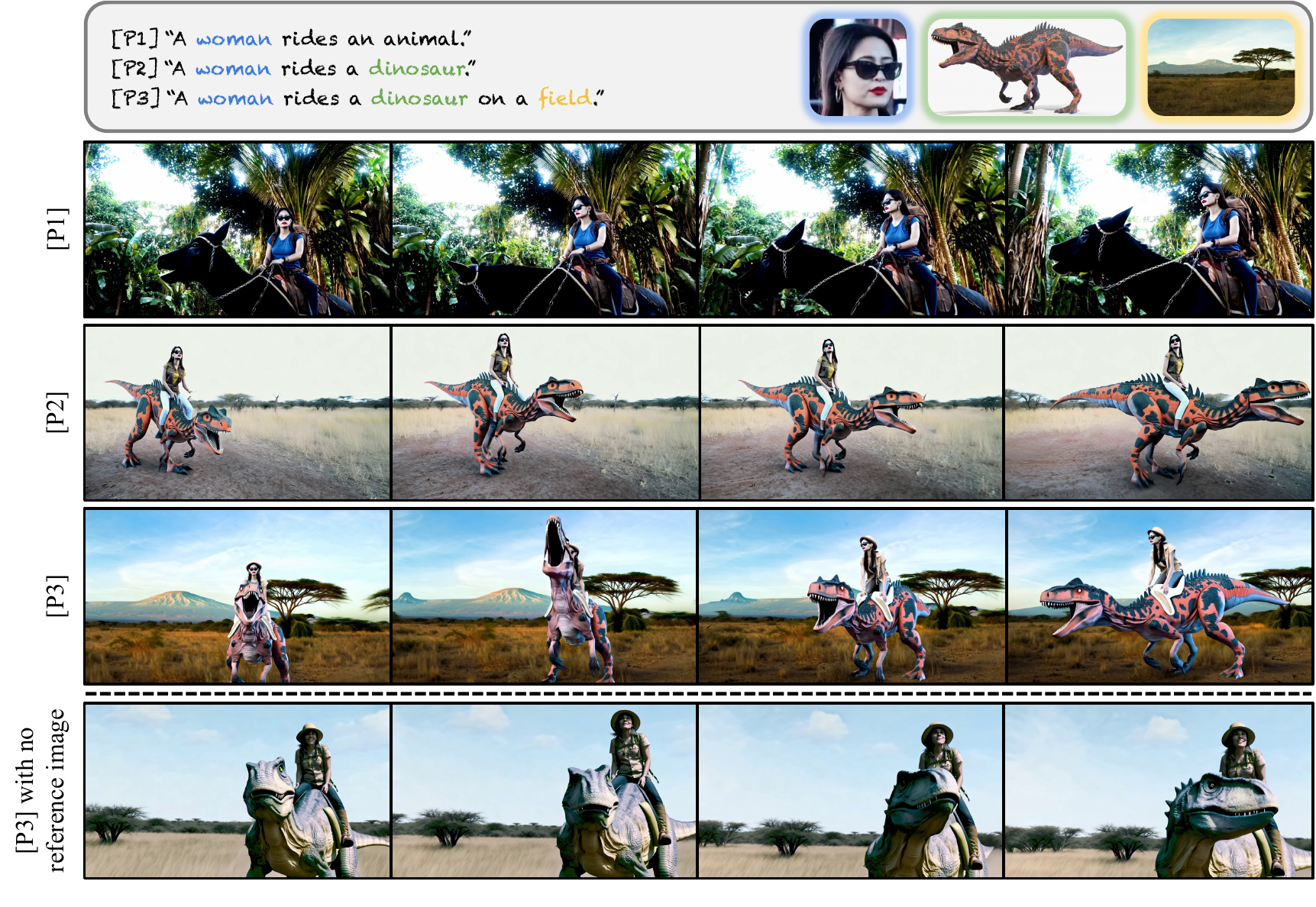}
    \mycaption{Additional results of multi-subject open-set personalization}{}
    \label{fig:multi_subject_dinosaur}
\end{figure*}

%% file: figures/multi_subject_discuss.tex
\begin{figure*}[t]
    \centering
    \includegraphics[trim={0 7mm 0 0}, width=.88\linewidth]{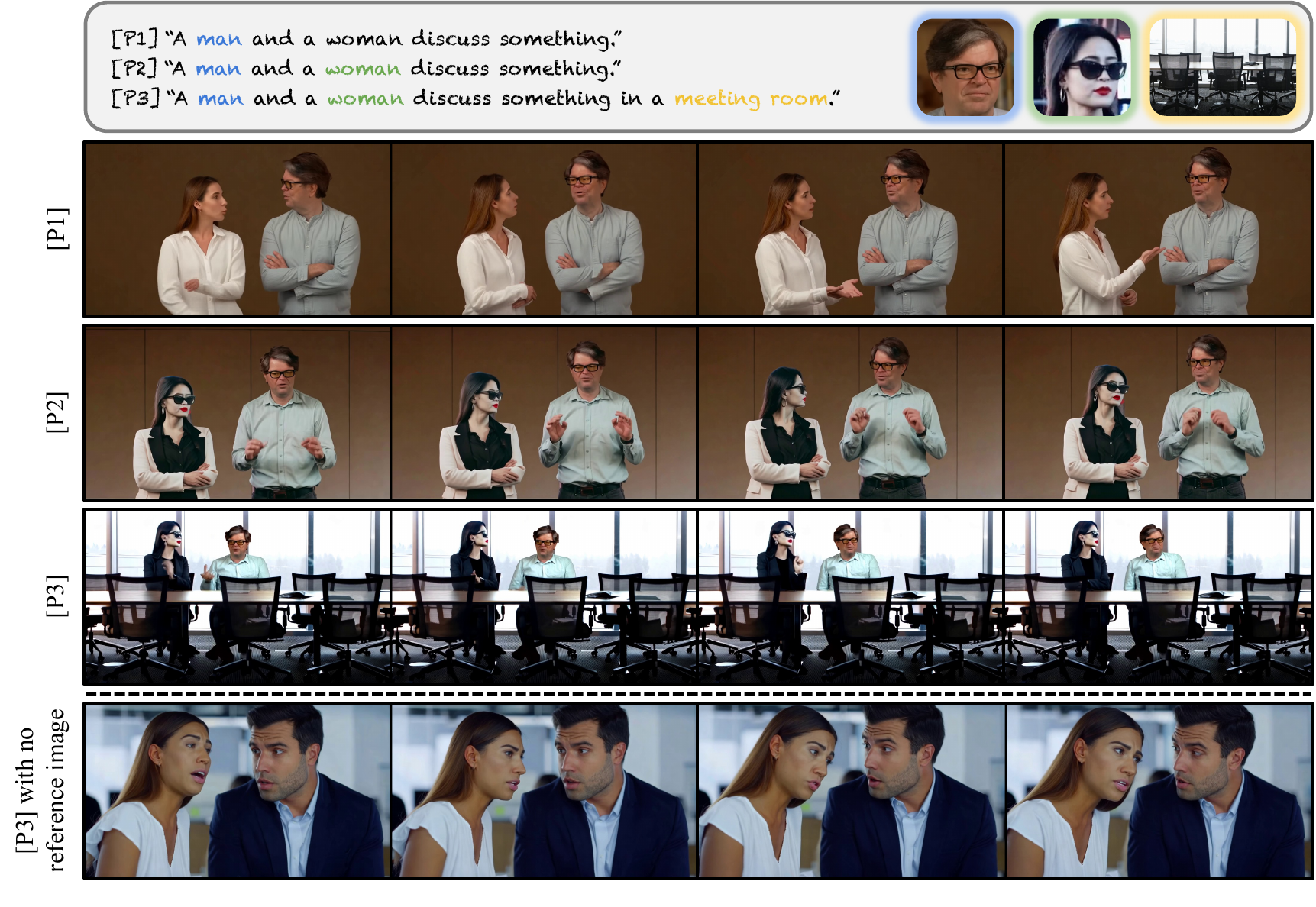}
    \mycaption{Additional results of multi-subject open-set personalization}{}
    \label{fig:multi_subject_discuss}
\end{figure*}

%% file: figures/multi_subject_drink_tea.tex
\begin{figure*}[t]
    \centering
    \includegraphics[trim={0 7mm 0 0}, width=.88\linewidth]{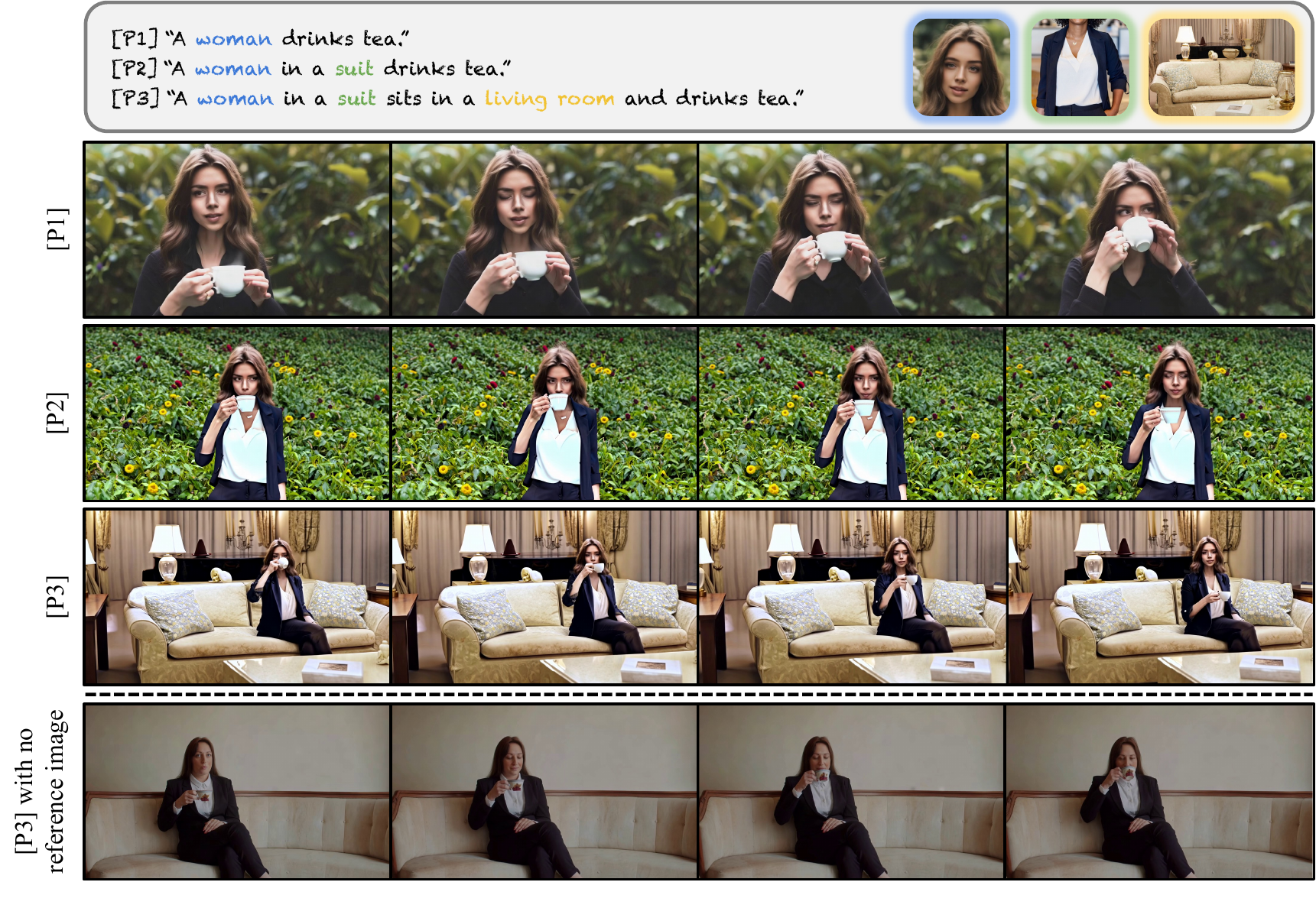}
    \mycaption{Additional results of multi-subject open-set personalization}{}
    \label{fig:multi_subject_drink_tea}
\end{figure*}

%% file: figures/multi_subject_rocket.tex
\begin{figure*}[t]
    \centering
    \includegraphics[trim={0 7mm 0 0}, width=.88\linewidth]{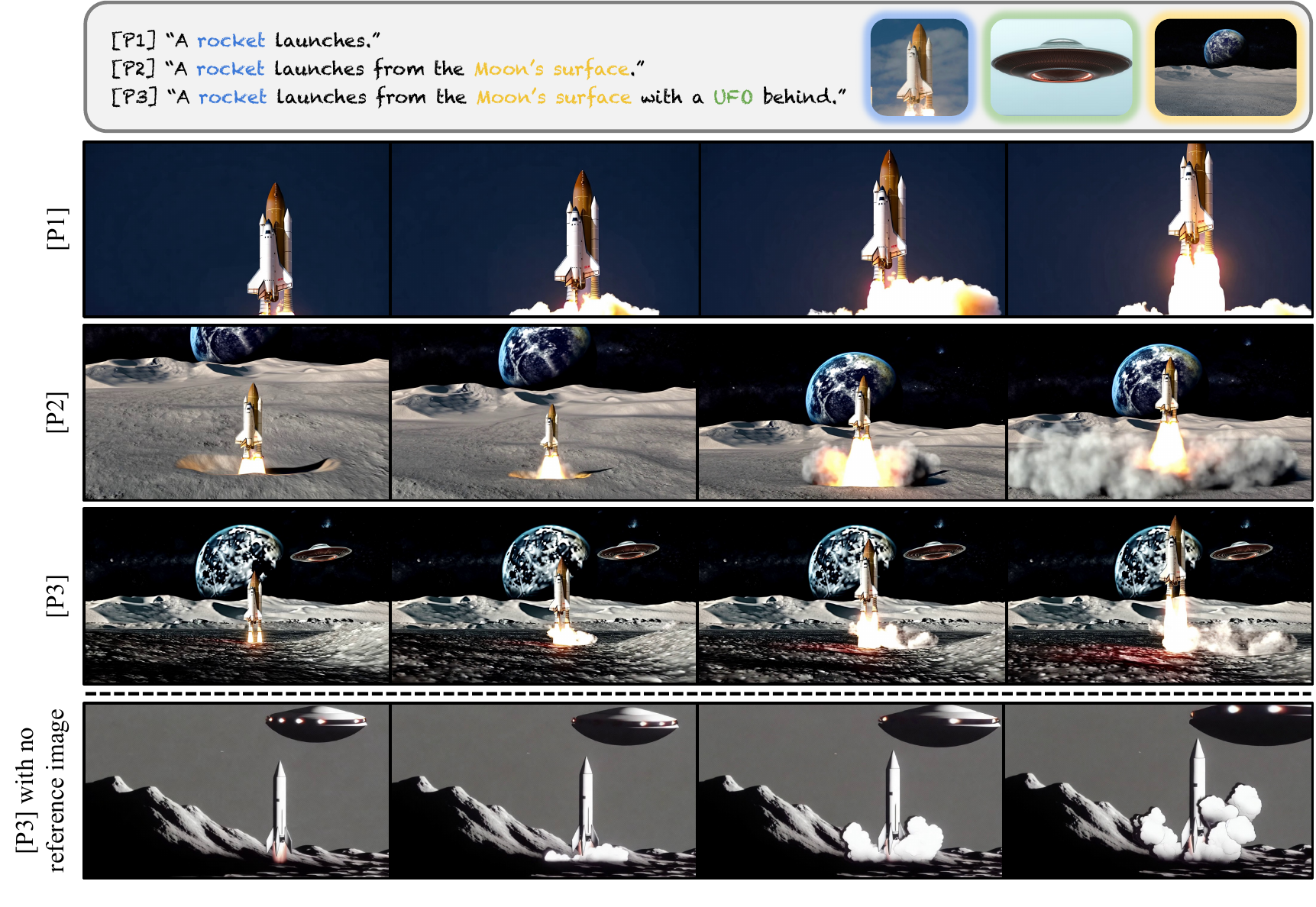}
    \mycaption{Additional results of multi-subject open-set personalization}{}
    \label{fig:multi_subject_rocket}
\end{figure*}

%% file: figures/ablate_person.tex
\begin{figure}[h]
    \centering
    \includegraphics[width=\linewidth]{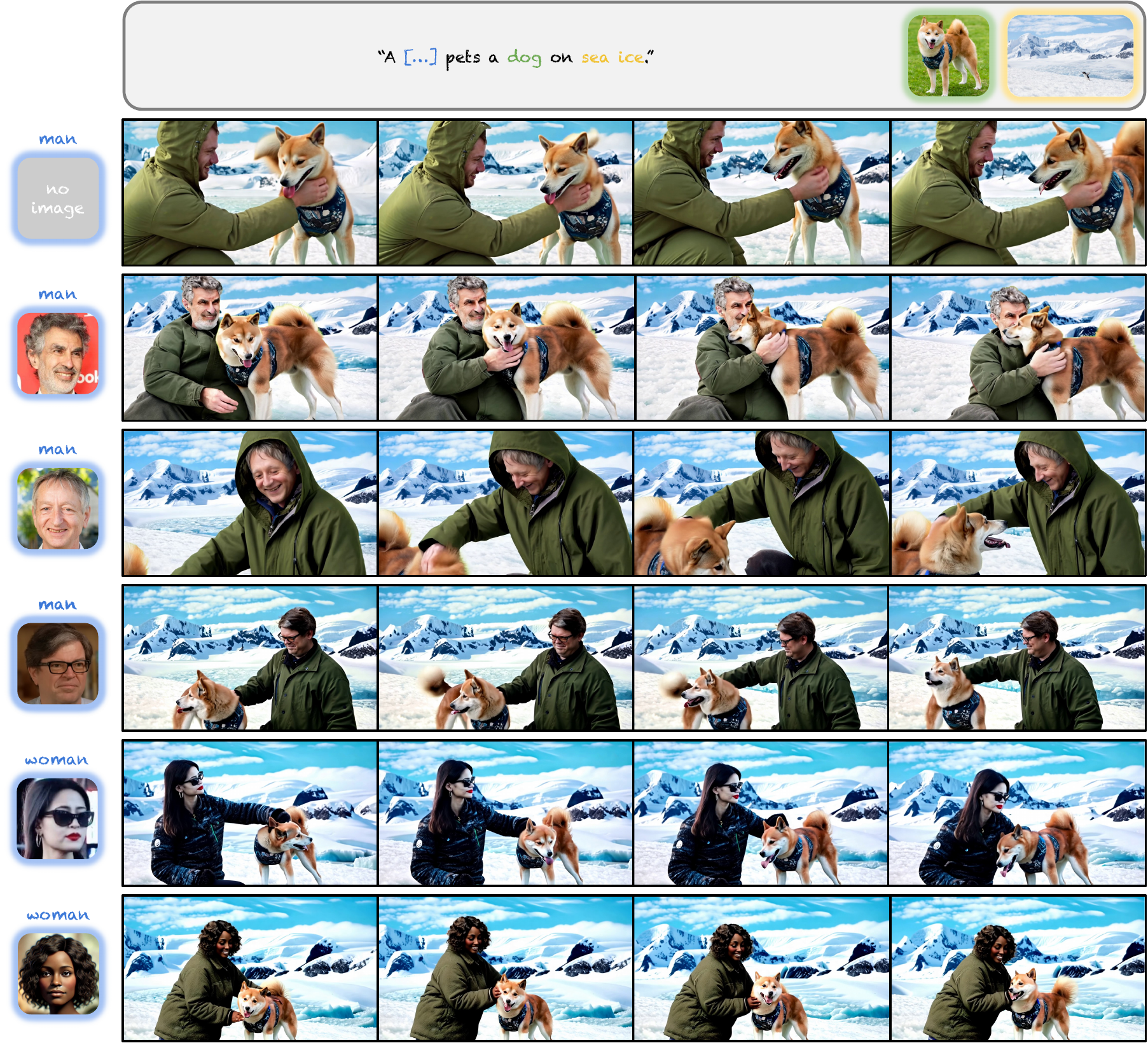}
    \vspace{-4mm}
    \mycaption{Same text prompt with different reference images of \textit{person}}{}
    \label{fig:ablate_person}
\end{figure}

%% file: figures/ablate_dog.tex
\begin{figure}[ht]
    \centering
    \includegraphics[width=\linewidth]{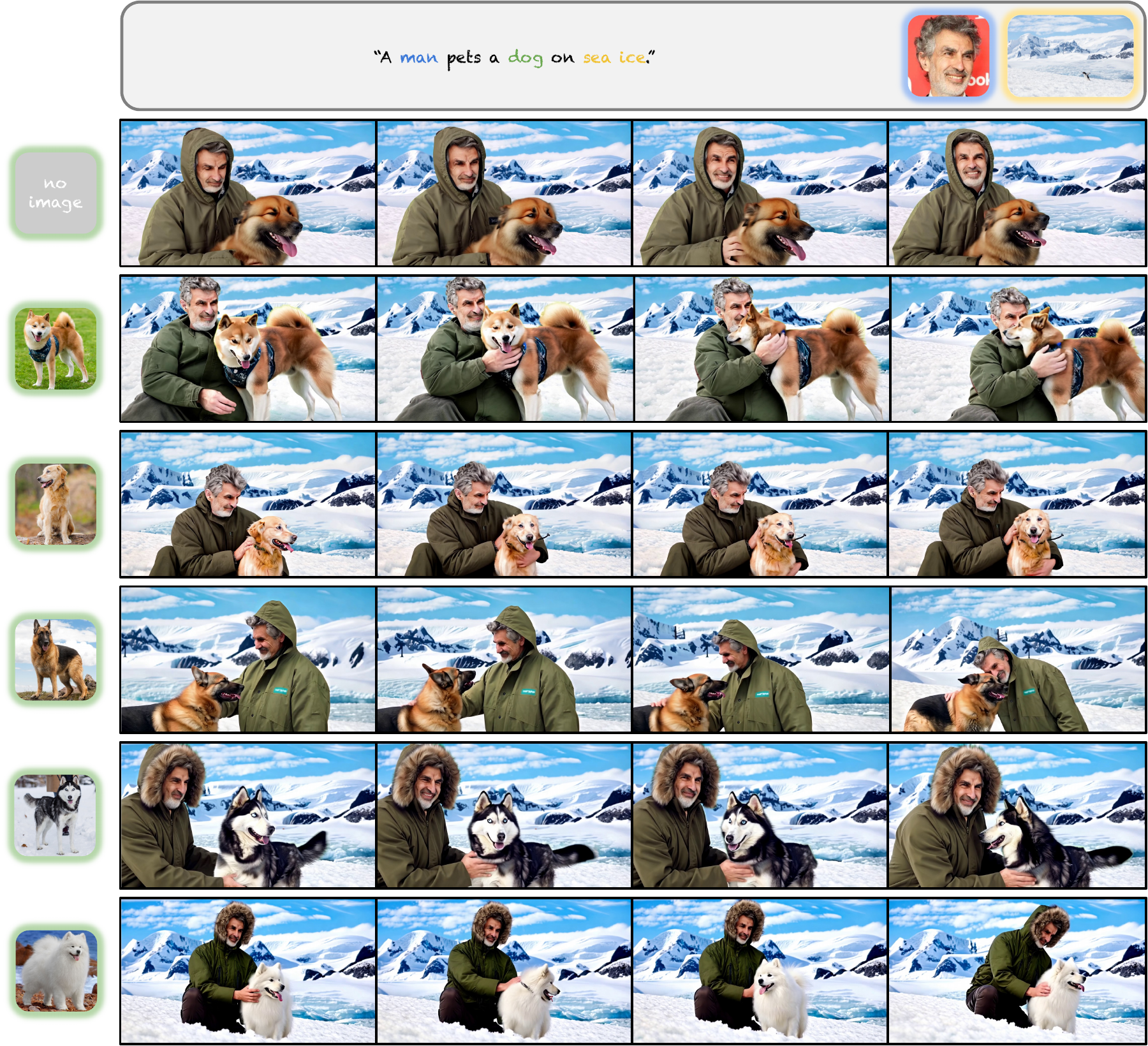}
    \vspace{-4mm}
    \mycaption{Same text prompt with different reference images of \textit{dog}}{}
    \label{fig:ablate_dog}
\end{figure}

%% file: figures/subject_mode_dog.tex
\begin{figure}[ht]
    \centering
    \includegraphics[width=.8\linewidth]{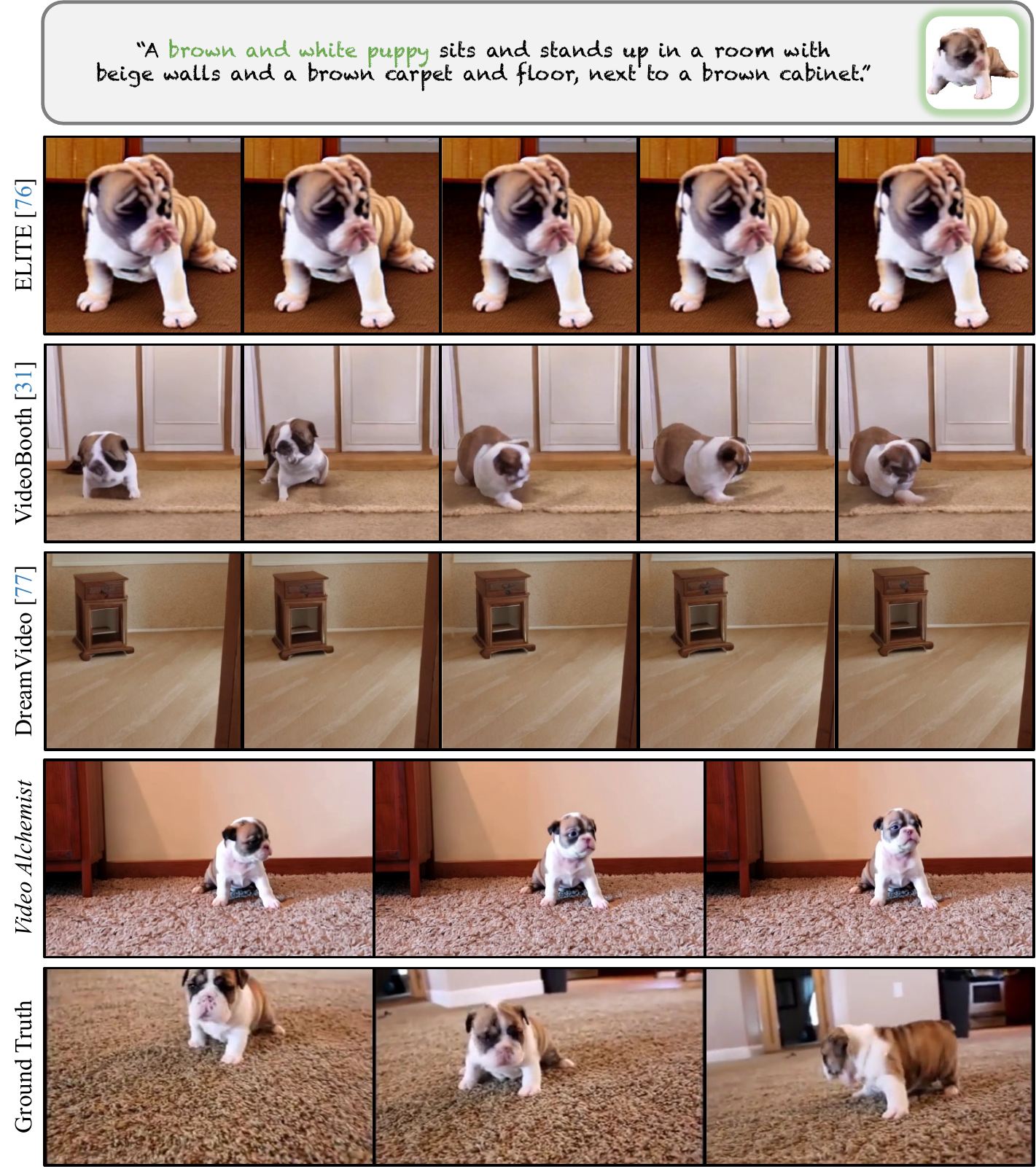}
    \mycaption{Qualitative comparison on the conditional subject of \textit{dog}}{}
    \label{fig:subject_mode_dog}
\end{figure}

%% file: figures/subject_mode_cat.tex
\begin{figure}[ht]
    \centering
    \includegraphics[width=.8\linewidth]{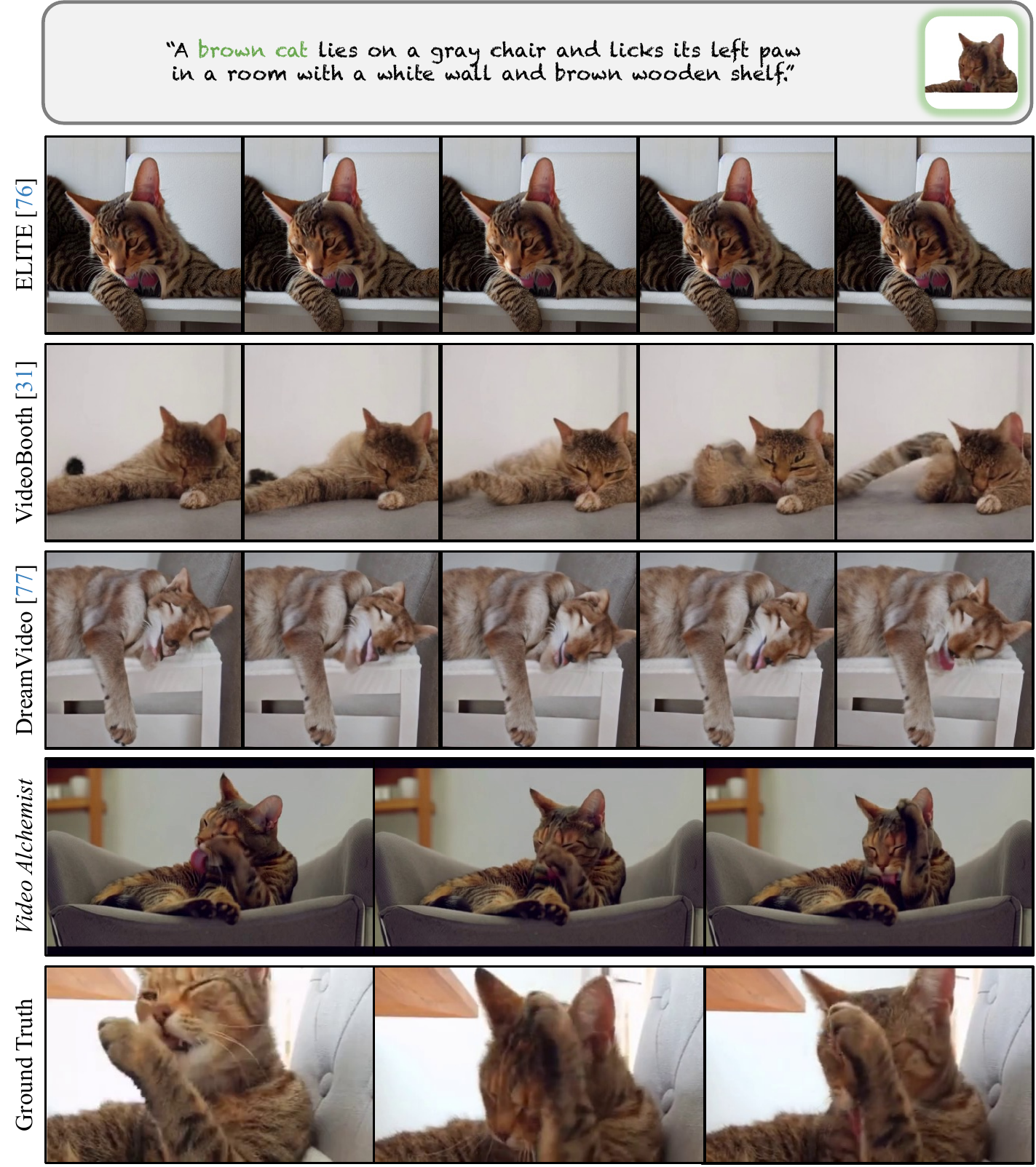}
    \mycaption{Qualitative comparison on the conditional subject of \textit{cat}}{}
    \label{fig:subject_mode_cat}
\end{figure}

%% file: figures/subject_mode_car.tex
\begin{figure}[ht]
    \centering
    \includegraphics[width=.8\linewidth]{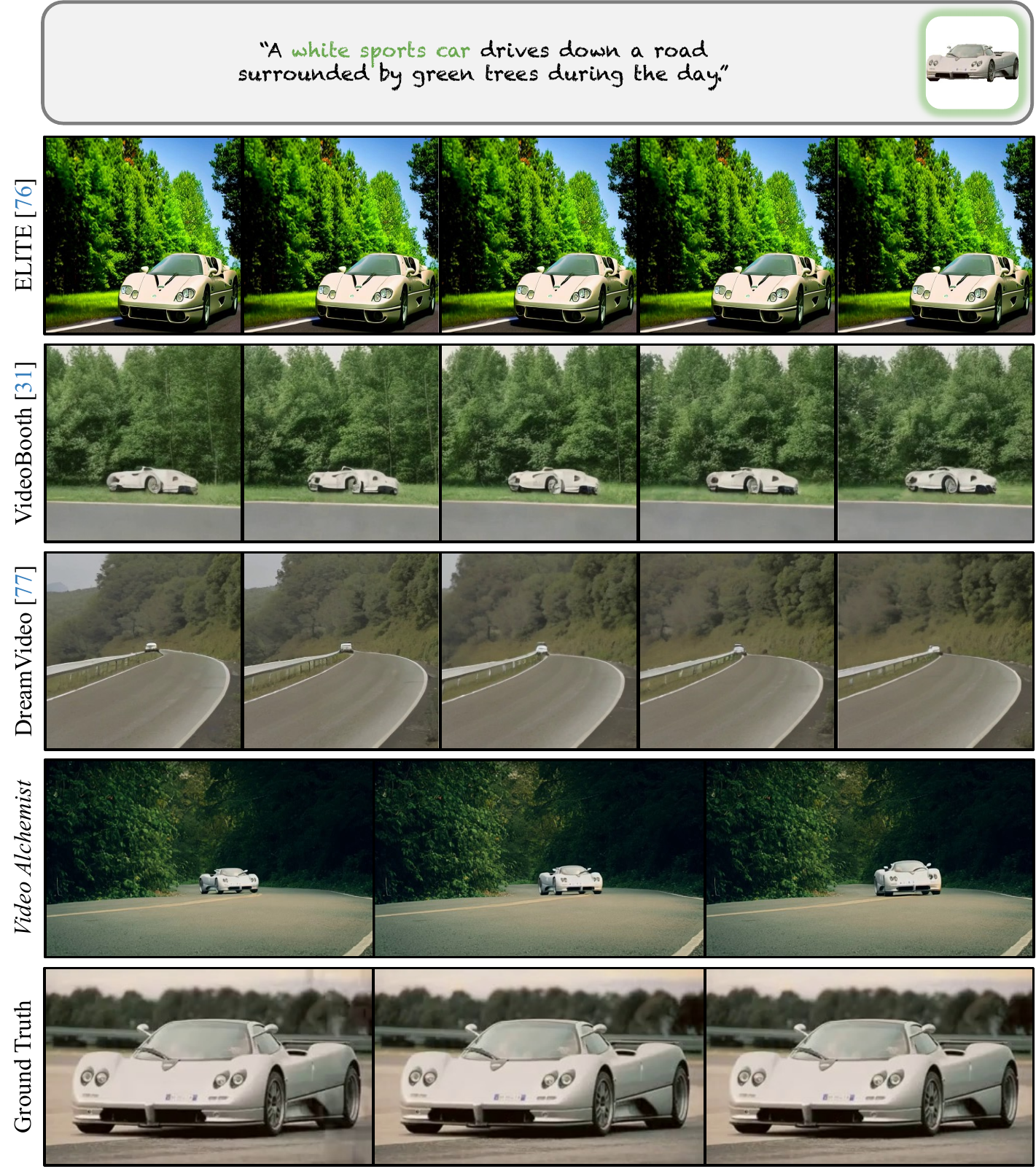}
    \mycaption{Qualitative comparison on the conditional subject of \textit{car}}{}
    \label{fig:subject_mode_car}
\end{figure}

%% file: figures/subject_mode_dinosaur.tex
\begin{figure}[ht]
    \centering
    \includegraphics[width=.8\linewidth]{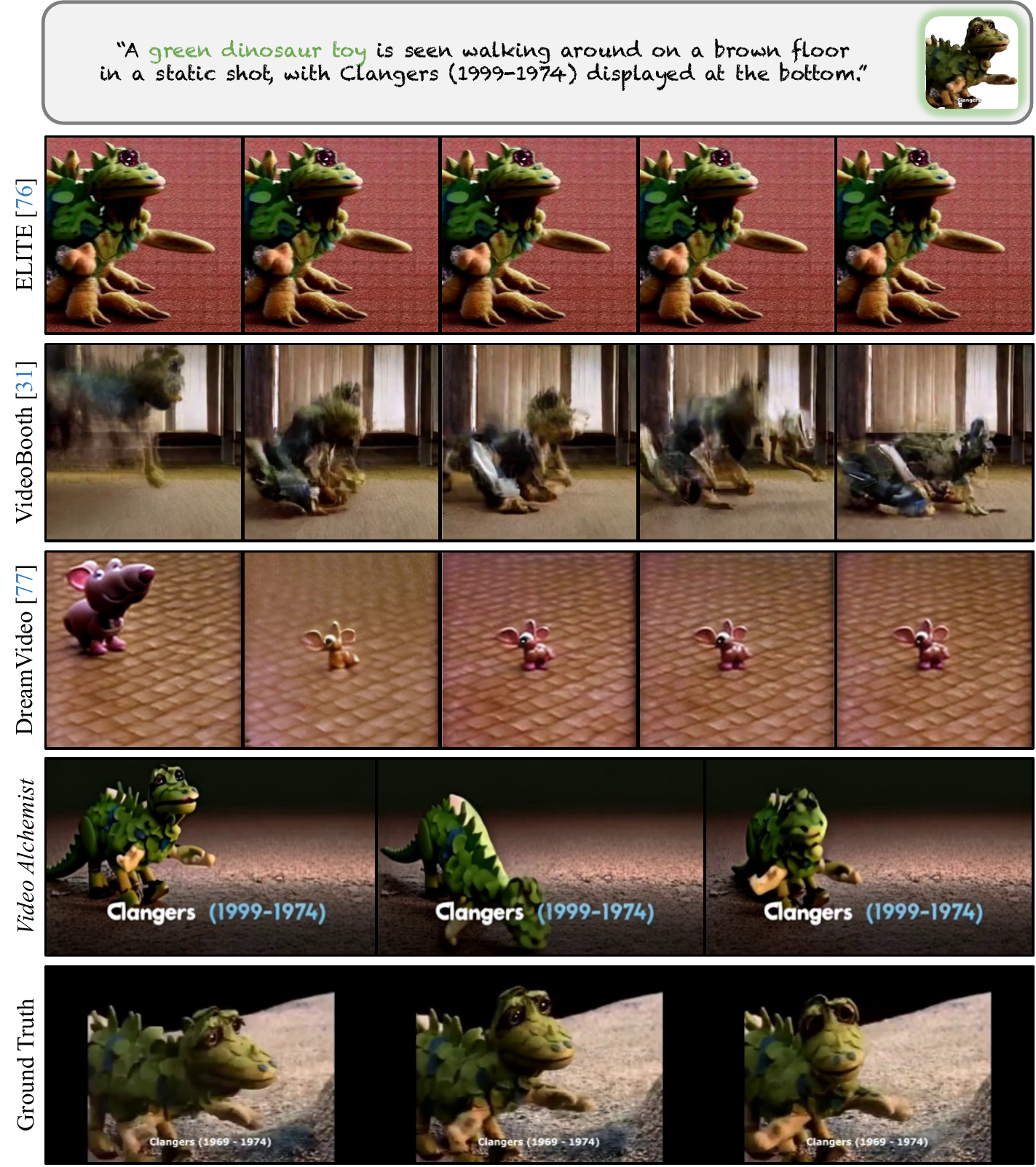}
    \mycaption{Qualitative comparison on the conditional subject of \textit{dinosaur toy}}{}
    \label{fig:subject_mode_dinosaur}
\end{figure}

%% file: sources/D_limitations.tex
\section{Limitations}
\label{app:limitation}

\myparagraph{Model Overfitting.}
In Section~\ref{sec:method_overfitting} and Appendix~\ref{app:details_dataset_augmentations}, we alleviate the model overfit by introducing data augmentation and random sampling with replacement during training. However, some undesirable image properties learned by the model remain unresolved. For example, \model may sometimes generate subjects with facial expressions or postures similar to the reference images. Figure~\ref{fig:msrvtt_comparison} shows that existing personalization models that adopt a similar reconstruction-based training, such as IP-Adapter~\cite{ip_adapter}, also exhibit the same problem, which remains a challenge for future work.

\vspace{2mm}
\myparagraph{Taking Image Segments as Inputs.}
Our model personalizes video synthesis using segmented images as input. Thus, additional user efforts may be required if localization algorithms are unable to segment the intended subject accurately. To address this problem, we plan to include training samples in which the segmented images are pasted onto random background images to ease the need to segment the reference images.

\vspace{2mm}
\myparagraph{Unnatural Composition for Multi-subject Conditioning.}
Empirically, for multi-subject conditioning, the synthetic videos sporadically exhibit unrealistic compositions and scales between different subjects. This behavior can be interpreted as the relative minority of videos with multiple subjects in the training dataset. We are considering creating a training dataset with a higher frequency of video samples with multiple subjects for future work.

\vspace{2mm}
\myparagraph{Unsupported Measure on Video Quality.}
Like the CLIP similarity score~\cite{torchmetrics}, \benchmark does not assess visual quality. Users must rely on alternative evaluations, such as user studies, to compare visual quality.